\documentclass{article}

\usepackage{microtype}
\usepackage{graphicx}
\usepackage{booktabs} 

\usepackage{hyperref}

\usepackage[accepted]{icml2025}
\usepackage{amsmath}
\usepackage{amssymb}
\usepackage{mathtools}
\usepackage{amsthm}
\usepackage{caption}
\usepackage{threeparttable}
\usepackage{xcolor, colortbl}
\usepackage{multirow}
\usepackage{enumitem}
\usepackage{wrapfig}
\usepackage{subcaption} 
\usepackage{xcolor}  

\newcommand{\best}[1]{{\textbf{\textcolor{red}{#1}}}}
\newcommand{\second}[1]{{\underline{\textcolor{blue}{#1}}}}
\usepackage[capitalize,noabbrev]{cleveref}

\theoremstyle{plain}

\theoremstyle{definition}

\theoremstyle{remark}

\usepackage{bbm}
\usepackage[textsize=tiny]{todonotes}

\begin{document}

\twocolumn[
\icmltitle{LLM-PS: Empowering Large Language Models for Time Series Forecasting \\with Temporal Patterns and Semantics}



\icmlsetsymbol{equal}{*}

\begin{icmlauthorlist}
	\icmlauthor{Jialiang Tang}{njust,ntu}
	\icmlauthor{Shuo Chen}{nju}
	\icmlauthor{Chen Gong}{sjtu}
	\icmlauthor{Jing Zhang}{whu}
	\icmlauthor{Dacheng Tao}{ntu}
\end{icmlauthorlist}

\icmlaffiliation{njust}{School of Computer Science and Engineering, Nanjing University of Science and Technology, China}
\icmlaffiliation{nju}{School of Intellgence Science and Technology, Nanjing University, China}
\icmlaffiliation{sjtu}{Department of Automation, Institute of Image ProceProceedingssing and Pattern Recognition, Shanghai Jiao Tong University, China}
\icmlaffiliation{whu}{School of Computer Science, Wuhan University, China}
\icmlaffiliation{ntu}{College of Computing and Data Science, Nanyang Technological University, Singapore}
\icmlcorrespondingauthor{Chen Gong}{chen.gong@sjtu.edu.cn}
\icmlcorrespondingauthor{Dacheng Tao}{dacheng.tao@ntu.edu.sg}

\icmlkeywords{Machine Learning, ICML}

\vskip 0.3in
]



\printAffiliationsAndNotice{}  

\begin{abstract}
Time Series Forecasting (TSF) is critical in many real-world domains like financial planning and health monitoring. Recent studies have revealed that Large Language Models (LLMs), with their powerful in-contextual modeling capabilities, hold significant potential for TSF. However, existing LLM-based methods usually perform suboptimally because they neglect the inherent characteristics of time series data. Unlike the textual data used in LLM pre-training, the time series data is semantically sparse and comprises distinctive temporal patterns. To address this problem, we propose LLM-PS to empower the LLM for TSF by learning the fundamental \textit{Patterns} and meaningful \textit{Semantics} from time series data. Our LLM-PS incorporates a new multi-scale convolutional neural network adept at capturing both short-term fluctuations and long-term trends within the time series. Meanwhile, we introduce a time-to-text module for extracting valuable semantics across continuous time intervals rather than isolated time points. By integrating these patterns and semantics, LLM-PS effectively models temporal dependencies, enabling a deep comprehension of time series and delivering accurate forecasts. Intensive experimental results demonstrate that LLM-PS achieves state-of-the-art performance in both short- and long-term forecasting tasks, as well as in few- and zero-shot settings.
\end{abstract}
\begin{figure}[t]
	\centering
	\includegraphics[scale=0.425]{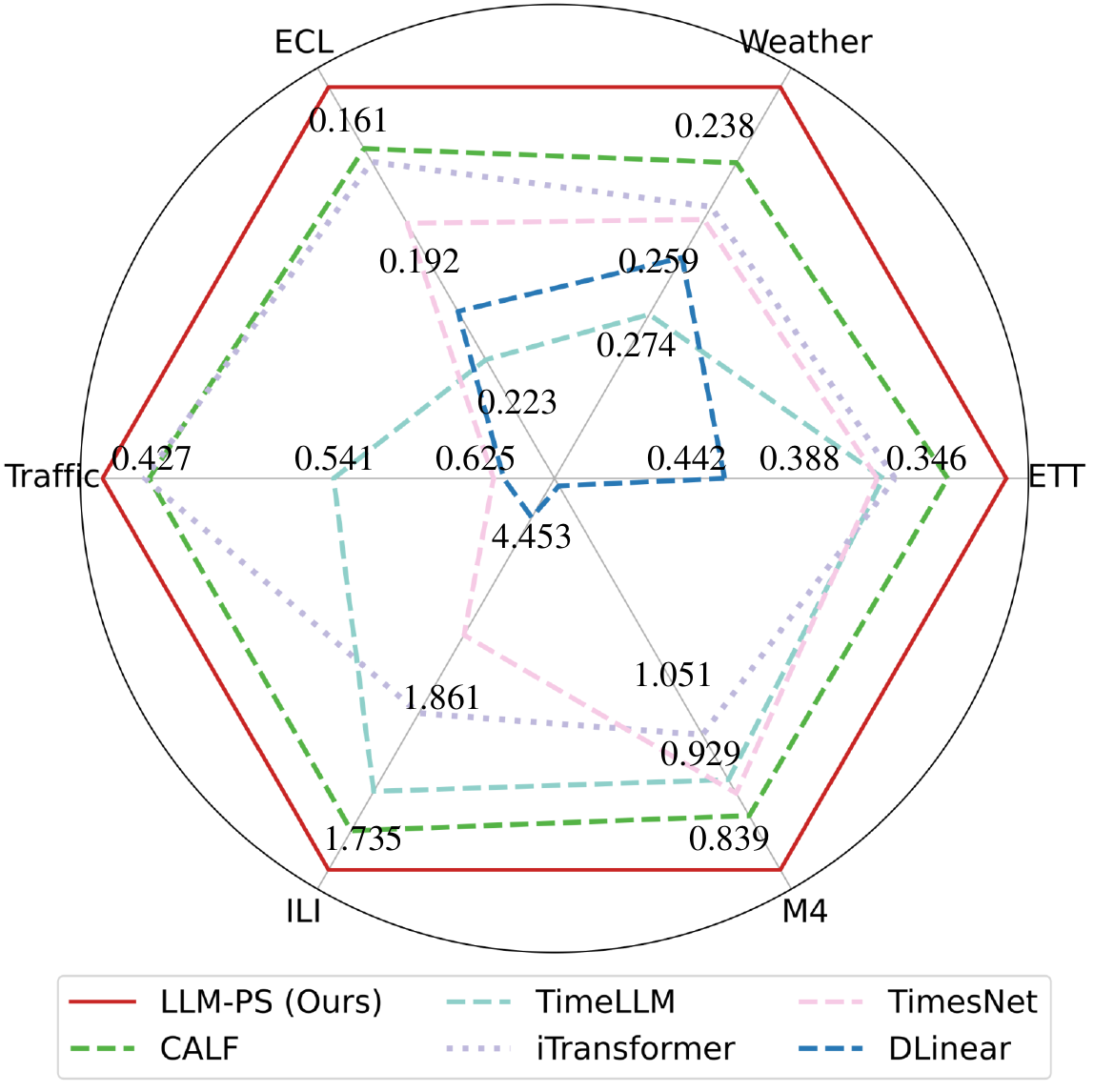}
	\caption{Performance of our proposed LLM-PS, LLM-based methods~\cite{liu2024taming,jintime}, and conventional deep learning methods~\cite{wu2023timesnet,zeng2023dlinear,liu2023itransformer}.}
	\label{fig_per}
\end{figure}
\section{Introduction}
\label{sec_intro}
Time Series Forecasting (TSF) plays a crucial role in various real-world applications, such as weather forecasting~\cite{angryk2020multivariate} and energy consumption prediction~\cite{electricity}. To achieve reliable TSF, traditional deep learning methods~\cite{wutimesnet,zhang2023crossformer,zhou2022fedformer} usually rely on domain-specific expertise to design customized models tailored to individual tasks. However, time series data often vary significantly across different domains~\cite{zhang2024large}. For example, stock prices in the financial market frequently fluctuate in short intervals, while temperature readings from the weather station typically evolve gradually over days~\cite{hyndman2018forecasting}, as shown in Fig.~\ref{fig_vistime}. Therefore, those task-specific models struggle to generalize effectively across different applications~\cite{jintime}. Moreover, these models are commonly trained from scratch and prone to overfitting in practical scenarios due to limited training data availability~\cite{wen2023transformers}.

Recently, Large Language Models (LLMs), such as GPT~\cite{brown2020language} and Llama~\cite{touvron2023llama}, have achieved great success in natural language processing. These LLMs are pre-trained on large-scale serialized textual datasets, endowing them with strong capabilities in both contextual modeling and generalization. In general, time series and textual data share two primary similarities: 1) \textit{Sequentiality}, both time series and textual data consist of ordered sets of elements; 2) \textit{Contextual dependency}, the meaning of a textual sentence relies on its context, and the value at the current time point is driven by its historical data. Building on these parallels, pioneering works~\cite{zhou2023one,liu2024taming,jintime,caotempo} attempt to fine-tune powerful LLMs for time series generation. Among them, Liu et al.~\cite{liu2024taming} align the distributions of time series and textual data to enhance the LLM's effectiveness in TSF. TimeLLM~\cite{jintime} bridges the modalities of time series and textual data by reprogramming time series data with text prototypes, thereby unlocking the TSF performance of LLMs.

Despite significant advancements in existing LLM-based methods for TSF, they generally prioritize aligning textual and time series data while ignoring the inherent characteristics of time series data, resulting in suboptimal performance. First, time series data exhibit \textit{diverse temporal patterns} rarely present in textual data, including regular short-term periodic fluctuations and persistent long-term trends that evolve over time~\cite{wu2021autoformer,zhou2022fedformer}. Second, the \textit{semantic information} is spare in time series data~\cite{cheng2023timemae}, usually requiring a prolonged period to convey specific semantics like ``rapid increase'' or ``sudden drop''. In comparison, words in textual data generally express explicit meaning, such as ``fast'' or ``slow''. Therefore, identifying fundamental temporal patterns and specific semantic information within time series data is crucial to guide LLMs for reliable time series prediction.

In this paper, we propose a novel LLM fine-tuning framework called LLM-PS, which enhances time series forecasting by leveraging the \textit{Patterns} and \textit{Semantics} in time series data. LLM-PS comprises two pivotal modules: the \textit{Multi-Scale Convolutional Neural Network} (MSCNN), designed to capture the intrinsic temporal patterns, and the \textit{Time-to-Text} semantic information extractor (T2T). Specifically, MSCNN extracts multi-scale features with varying receptive fields by hierarchical stacked convolutional layers. Features with small receptive fields primarily capture short-term patterns (\textit{i.e.}, periodicity fluctuations), while those with large receptive fields focus on long-term patterns (\textit{i.e.}, global trends). To further cope with short-term and long-term patterns, we decouple them from multi-scale features based on the wavelet transform and enhance them via global-to-local and local-to-global assembling. Meanwhile, T2T extracts valuable semantics by precisely predicting labels for time-series patches under a high masking ratio. Subsequently, the temporal patterns and semantics are integrated through feature transfer and input into the LLM to generate time series. Thanks to its ability to effectively handle diverse temporal patterns and accurately extract semantically rich information, LLM-PS comprehensively understands time series data, consistently achieving state-of-the-art (SOTA) performance across multiple datasets, as shown in Fig.~\ref{fig_per}.
\begin{itemize}
	\itemsep=1pt
	\item We recognize that time series data exhibits intrinsic characteristics that differ from those in textual data used in LLM pre-training and propose a novel TSF framework to leverage these distinctive properties to derive the LLM for reliable time series forecasting.
	\item We design new MSCNN and T2T modules tailored for effectively handling the diverse temporal patterns and accurately extracting semantic information, and thus enhancing the LLM's understanding of the input time series during forecasting.
	\item Our LLM-PS consistently achieves SOTA performance across a variety of mainstream time series prediction tasks, especially in few-shot and zero-shot scenarios. Furthermore, our model is highly efficient while robust to noise compared with other popular methods.
\end{itemize}

\section{Related Works}
\label{sec_realted}
\subsection{Time Series Forecasting}
Time series forecasting aims to predict future values of a series based on historical data, serving as a crucial capability in industries like finance management~\cite{patton2013copula}, weather forecasting~\cite{angryk2020multivariate}, and energy consumption prediction~\cite{haoyietalinformer}. Recently, TSF methods have evolved from traditional statistical models~\cite{taylor2018forecasting,oreshkin2019n} to sophisticated deep learning models~\cite{wutimesnet,wangtimemixer}. Deep learning methods generally utilize Recurrent Neural Networks (RNNs)~\cite{salinas2020deepar}, Convolutional Neural Networks (CNNs)~\cite{wutimesnet}, Transformers~\cite{wu2021autoformer}, and Multi-Layer Perceptrons (MLPs)~\cite{wangtimemixer} as their backbones. By leveraging domain expertise, they can perform well on specific tasks. Nonetheless, their real-world applicability is usually constrained by the variability of temporal patterns across domains~\cite{jintime}.

To address these challenges, some researchers~\cite{zhou2023one,liu2024taming} have turned to LLMs for TSF and achieved great success. Current methods primarily focus on bridging the gap between time series and textual modalities. For example, PromptCast~\cite{xue2023promptcast} encodes time series and textual data into prompts to guide the LLM prediction. The prompts contain contextual information, task requirements, and the desired output format. TimeLLM~\cite{jintime} further enhances guidance for LLM by incorporating domain information, instructions, and data statistics within the prompts. Instead of designing prompts,~\cite{chung2023text} directly train a codebook to convert continuous time series into discrete input embeddings by mapping them to the most similar codewords in the codebook. Similarly, ~\cite{rubenstein2023audiopalm} apply K-Means clustering to time series embeddings and construct a codebook~\cite{van2017neural} by the cluster centroids. Additionally, CALF~\cite{liu2024taming} trains separate LLM branches for time series and textual data, which aligns their features across the intermediate and output layers. LLM-TS~\cite{chen2024llm} employs a CNN as the time series branch and guides it with the LLM by minimizing their mutual information. 

Despite the promising performance of LLM-based methods, they inadequately solve the intrinsic characteristics of time series data, limiting their effectiveness for TSF. Our LLM-PS effectively handles these properties of temporal patterns and semantics, thereby achieving reliable performance.
\begin{figure*}[t]
	\centering
	\includegraphics[scale=0.5]{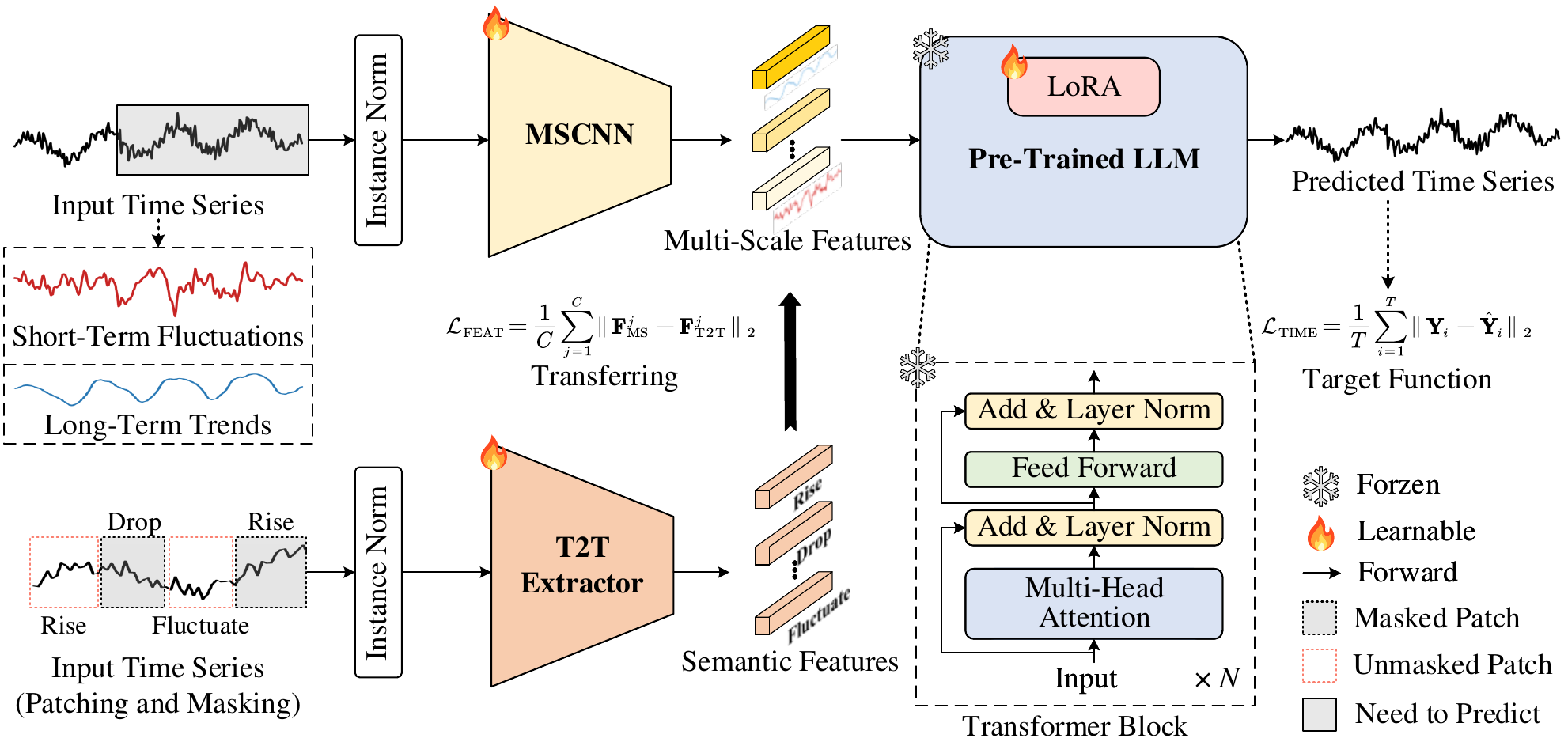}
	\caption{An overview of our proposed LLM-PS. Our LLM-PS incorporates a Multi-Scale Convolutional Neural Network (MSCNN) and Time-to-Text (T2T) semantics extractor. Specifically, for input time series $\mathbf{Y}$, MSCNN constructs multi-scale features $\mathbf{F}_{\text{MS}}$ with various receptive fields (darker colors indicate larger receptive fields), thereby capturing localized short-term fluctuations and broader long-term trends. T2T extracts features $\mathbf{F}_{\text{T2T}}$ with meaningful semantics to promote the LLM to precisely understand the input time series. Finally, the diverse temporal patterns and rich semantics are integrated via feature transferring and input into the LLM to generate precise time series $\hat{\mathbf{Y}}$.}
	\label{fig_structure}
\end{figure*}
\subsection{Temporal Patterns Learning}
Time series data comprises short-term fluctuations and long-term trends~\cite{zhang2005neural}. To capture these temporal patterns, existing methods~\cite{wangtimemixer,kowsher2024llm} convert the original inputs to multiple time series inputs with varying scales through pooling operations with various window sizes. Consequently, the model can learn short-term periods and long-term trends from low-scale and high-scale inputs, respectively. However, these methods incur prohibitive computational overheads due to the intricate processing of numerous scaled signals. To address this, we introduce a new MSCNN that enables efficiently generating multi-scale features with diverse temporal patterns using a single forward, as analyzed in Section~\ref{sec_exp_ma}.

Besides to construct multi-scale signals, some methods~\cite{wu2021autoformer,wu2023interpretable} directly separate temporal patterns from time series. Earlier studies~\cite{wu2021autoformer,wu2023interpretable} employ average pooling to extract long-term trends from time series data while considering the remaining segments as short-term patterns. Recent studies~\cite{zhou2022fedformer,wangtimemixer} highlighted that low-frequency and high-frequency components in the frequency domain correspond to long-term and short-term patterns, respectively. Therefore, they utilize the Fourier transform~\cite{cochran1967fast} to isolate them. In this paper, we decouple short-term and long-term patterns based on the wavelet transform~\cite{zhang2019wavelet}. Unlike the aforementioned methods that rely on average pooling in the temporal domain or Fourier transform in the frequency domain, our method concurrently learns from both the temporal and frequency domains. Therefore, our approach can accurately decompose short-term and long-term components, as visualized in Fig.~\ref{fig_visdtp}.
\section{Approach}
\label{sec_app}
Time series forecasting aims to predict the series $\mathbf{Y}\in \mathbb{R}^{T \times V}$ for the next $T$ time steps given the $H$ time steps historical observations $\mathbf{X}\in \mathbb{R}^{H \times V}$, where $V$ denotes the number of variables. As shown in Fig.~\ref{fig_structure}, our LLM-PS incorporates a new MSCNN (detailed in Sections~\ref{sec_app_msc}\&\ref{sec_app_dm}) to learn multi-scale features from the input time series, which captures both short-term and long-term patterns. Meanwhile, within LLM-PS, the T2T module (described in Section~\ref{sec_app_t2t}) enriches the multi-scale features by the semantic information extracted from the input time series. Consequently, the multi-scale features with diverse temporal dependencies and valuable semantics are input into the LLM to facilitate accurate future time series $\hat{\mathbf{Y}} \in \mathbb{R}^{T \times V}$.
\begin{figure}[t]
	\centering
	\includegraphics[scale=0.47]{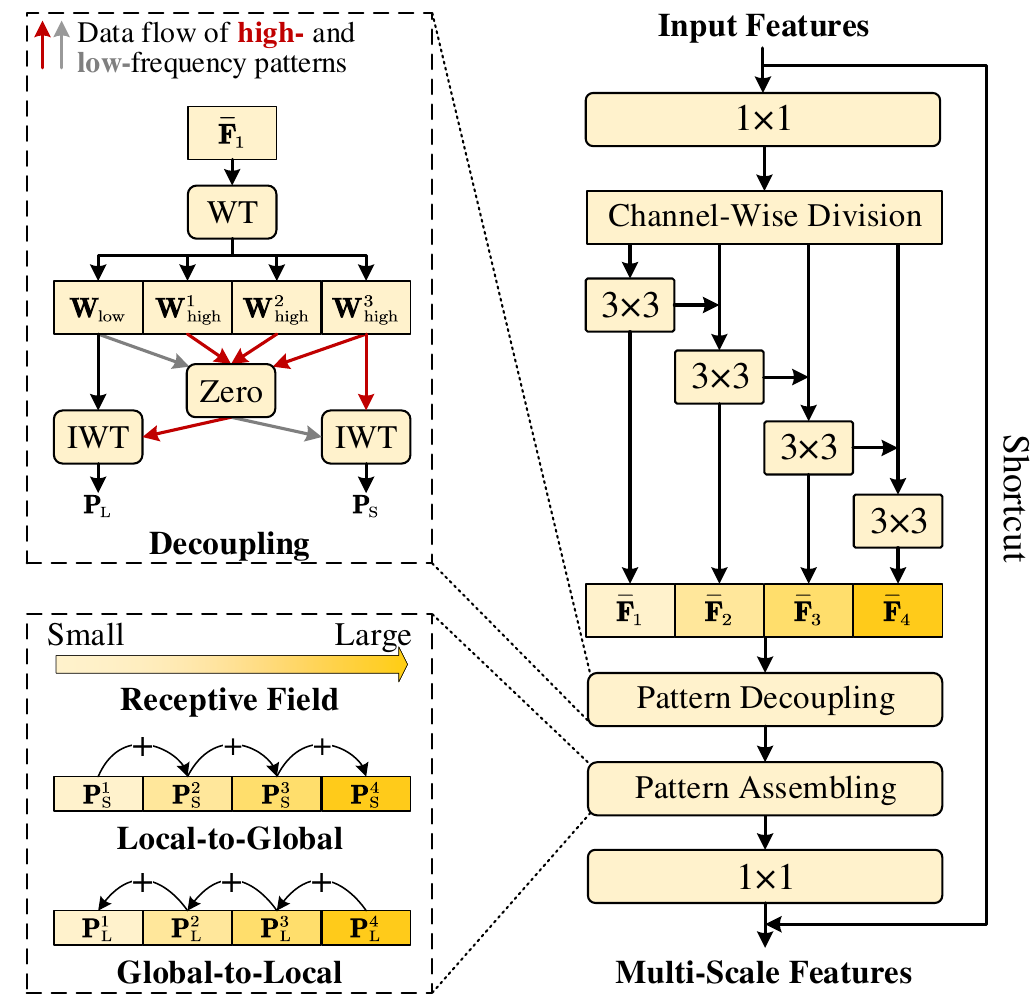}
	\caption{The diagram of our MSCNN block. The divided features are initially fed into their related 3$\times$3 convolutional layers to obtain features (\textit{e.g.}, $\bar{\mathbf{F}}_{1}$) with various receptive fields. Then, these features are decoupled into long-term patterns (\textit{e.g.}, $\mathbf{P}^{1}_{\text{L}}$) and short-term patterns (\textit{e.g.}, $\mathbf{P}^{1}_{\text{S}}$) using the Wavelet Transform (WT) and Inverse Wavelet Transform (IWT). Subsequently, the long-term and short-term patterns are enhanced through global-to-local and local-to-global assembling, respectively. Finally, the improved patterns are added together and passed through a 1$\times$1 convolutional layer to obtain the multi-scale features.}
	\label{fig_mscnn}
\end{figure}
\subsection{Multi-Scale Convolutional Neural Network}
\label{sec_app_msc}
Real-world time series data is inherently complex, manifesting short-term and long-term patterns~\cite{cleveland1990stl,zhang2005neural}, both of which are critical for accurate forecasting. Short-term patterns reflect localized fluctuations and periodic dynamics, while long-term patterns encapsulate broader trends that signal future trajectories. In conventional CNNs, each convolutional layer has a fixed receptive field, which limits its output features to a narrow temporal scope with only a single temporal pattern. To capture diverse temporal patterns, we follow classic CNNs~\cite{he2016deep,gao2019res2net} and design a new MSCNN stacked with bottleneck blocks. As depicted in Fig.~\ref{fig_mscnn}, each MSCNN block learns multi-scale features by parallel branches. Features with smaller receptive fields focus on periodic fluctuations, whereas those with larger receptive fields concentrate on overarching trends.

In each MSCNN block, the $C$ channels input features $\mathbf{F}_{\text{in}}\in \mathbb{R}^{C \times V}$ first undergo 1$\times$1 convolutional layer before partitioned to $B$ branches $\{\mathbf{F}_{1}, \ldots, \mathbf{F}_{B}\}$, where $\mathbf{F}_{i}\in \mathbb{R}^{C/B \times V}$ (batch dimension is omitted for simplify description). Then, $B$ branches features are recursively fed into their respective 3$\times$3 convolutional layer and added with the output of the preceding branches (except for $\mathbf{F}_{1}$), as follows:
\begin{equation}
\bar{\mathbf{F}}_i= \begin{cases}
\mathbf{Conv}_{i}\left(\mathbf{F}_i\right), & i=1, 
\\ \mathbf{Conv}_i\left(\mathbf{F}_i+\bar{\mathbf{F}}_{i-1}\right), & 1<i \leq B.\end{cases}
\label{eq_mscnn}
\end{equation}
The receptive fields of features in $\{\bar{\mathbf{F}}_{1}, \ldots, \bar{\mathbf{F}}_{B}\}$ increase sequentially as each $\bar{\mathbf{F}}_{i}$ (if $i>1$) aggregates information from preceding branches (detailed in \textbf{Appendix}~\ref{sec_apx_rf}). Finally, these features are concatenated together and fused through a 1$\times$1 convolution layer to derive the output features $\mathbf{F}_{\text{out}}$:
\begin{equation}
\mathbf{F}_{\text{out}}=\text{Conv}_{1\times 1}(\text{Concate}(\{\bar{\mathbf{F}}_{1}, \ldots, \bar{\mathbf{F}}_{B}\}) + \mathbf{F}_{\text{in}},
\end{equation}
where $\mathbf{F}_{\text{in}}$ is added in the output features via shortcut. Afterward, $\mathbf{F}_{\text{out}}$ is input into the subsequent MSCNN blocks to produce the multi-scale features $\mathbf{F}_{\text{MS}}$ for the LLM.
\subsection{Temporal Patterns Decoupling and Assembling}
\label{sec_app_dm}
In the previous section, we show how MSCNN captures diverse temporal patterns by learning multi-scale features. For a time series, its high-frequency and low-frequency components can effectively represent the corresponding short-term and long-term temporal patterns, respectively~\cite{wangtimemixer,kowsher2024llm}. Therefore, we introduce a novel patterns decoupling-assembling mechanism based on the wavelet transform~\cite{zhang2019wavelet} (as shown in Fig.~\ref{fig_mscnn}) to further refine temporal patterns in the multi-scale features.

Initially, the features $\left\{\bar{\mathbf{F}}_{1}, \ldots, \bar{\mathbf{F}}_{B}\right\}$ output by $B$ branches in the MSCNN block are decoupled into low-frequency component $\mathbf{W}_{\text{low}}$ and high-frequency component $\mathbf{W}_{\text{high}}$ using the wavelet transform (see \textbf{Appendix}~\ref{sec_apx_wt}) $\text{WT}(\cdot)$:
\begin{equation}
\mathbf{W}^{b}_{\text{low}}, \{\mathbf{W}^{b}_{\text{high}\_i}\}_{i=1}^{w}\!=\!\text{WT}(\bar{\mathbf{F}}_{b}, w), \quad \!\!b\in \{1, \ldots, B\}\!,
\end{equation}
where $w$ denotes the decomposition levels. Subsequently, the short-term pattern $\mathbf{P}_{\text{S}}$ and long-term pattern $\mathbf{P}_{\text{L}}$ are constructed by the inverse wavelet transform $\text{IWT}(\cdot)$:
\begin{equation}
\left\{
\begin{aligned}
\mathbf{P}_{\text{S}}^{b} &= \text{IWT}(\text{Zero}(\mathbf{W}^{b}_{\text{low}}), \{\mathbf{W}^{b}_{\text{high}\_i}\}_{i=1}^{w}), \\
\mathbf{P}_{\text{L}}^{b} &= \text{IWT}(\mathbf{W}^{b}_{\text{low}}, \{\text{Zero}(\mathbf{W}^{b}_{\text{high}\_i})\}_{i=1}^{w}),
\end{aligned}
\right.
\end{equation}
where $\text{Zero}(\cdot)$ operation generates features matching the input dimensions but filled with zeros.

For features in $\left\{\bar{\mathbf{F}}_{1}, \ldots, \bar{\mathbf{F}}_{B}\right\}$, their receptive fields expand sequentially. Features with smaller receptive fields more effectively capture local periodic fluctuations, whereas those with larger receptive fields focus on broader global trends. Therefore, short-/long-term patterns $\mathbf{P}_{\text{S}}$/$\mathbf{P}_{\text{L}}$ are reinforced in local-to-global/global-to-local assembling, as follows:
\begin{equation}
\left\{
\begin{aligned}
&\text { For } b: 2 \rightarrow B \text { do: } &\quad \mathbf{P}_{\text{S}}^{b}=\mathbf{P}_{\text{S}}^{b} + \mathbf{P}_{\text{S}}^{b-1}, \\
&\text { For } b: (B-1) \rightarrow 1 \text { do: } &\quad \mathbf{P}_{\text{L}}^{b}=\mathbf{P}_{\text{L}}^{b} + \mathbf{P}_{\text{L}}^{b+1}.
\end{aligned}
\right.
\end{equation}
After the assembling, features in $\left\{\bar{\mathbf{F}}_{1}, \ldots, \bar{\mathbf{F}}_{B}\right\}$ are reconstructed by combining the short-term and long-term patterns:
\begin{equation}
\bar{\mathbf{F}}_{b}=\mathbf{P}_{\text{S}}^{b} + \mathbf{P}_{\text{L}}^{b}, \quad b\in \{1, \ldots, B\}.
\end{equation}
\subsection{Time-to-Text Semantics Extraction}
\label{sec_app_t2t}
LLMs are pre-trained on extensive textual data rich in semantic information, where individual words carry clear meaning. In comparison, time series data is semantically sparse, requiring an entire sequence to convey specific content. Consequently, LLMs pre-trained on textual data struggle to precisely interpret semantics present in time series data. Recent advancements in audio processing~\cite{defossezhigh,zhang2024speech} suggest that self-supervised learning~\cite{bert} can promote models to understand the serialized audio sequences that are similar to time series data. Inspired by them, we propose the T2T module with an encoder-decoder structure to extract semantics within time series data for the LLM, as shown in Fig.~\ref{fig_t2t}.

Initially, the input time series $\mathbf{X}\in \mathbb{R}^{H \times V}$ is divided into $P$ patches $\{\mathbf{X}_{i}\}^{P}_{i=1}$, where $\mathbf{X}_{i}\in \mathbb{R}^{L \times V}$ and $L$ denotes the patch length. Following~\cite{hsu2021hubert}, approximately 75\% of the patches are randomly masked. During training, the T2T module learns precise semantic information by reconstructing the masked patches while predicting their labels. The loss function is defined as:
\begin{equation}
\mathcal{L}_{\text{T2T}}=\frac{1}{P}\sum_{i=1}^P(\mathbbm{1}_{[\text{M}(i)=1]}\|\mathbf{X}_{i}-\hat{\mathbf{X}}_{i}\|_2+ l_{i}\log\frac{l_{i}}{\hat{l_{i}}}).
\end{equation}
Here, $\hat{\mathbf{X}}_{i}$ represents the reconstructed patch,  $\mathbbm{1}_{[\text{M}(i)=1]}$ is the indicator function that equals 1 if the $i$-th patch is masked (\textit{i.e.}, $\text{M}(i)$=1); $l_{i}$ and $\hat{l}_{i}$ denote the semantic labels of the original and reconstructed patches, respectively. For a patch $\mathbf{X}_{i}$ (resp. $\hat{\mathbf{X}}_{i}$), its semantic label $l_{i}$ (resp. $\hat{l}_{i}$) is assigned as the word with the most similar LLM text embeddings, based on the similarity $\mathbf{S}_{i}$ calculated as: 
\begin{equation}
\mathbf{S}_{i} = \text{Proj}(\mathbf{X}_{i}) \cdot \mathbf{E}^\top.
\end{equation}
Here, $\text{Proj}(\cdot)$ linearly transforms the input to the same dimensions of the LLM text embeddings $\mathbf{E}$, and ``$\top$" denotes the transpose. Further details of T2T and the semantic label assignment process are provided in the \textbf{Appendix}~\ref{sec_apx_t2t}.
\subsection{Efficient Training of LLM-PS}
To efficiently train the LLM with huge parameters, we employ the parameter-efficient  Low-Rank Adaptation (LoRA)~\cite{hu2021lora} to fine-tune the LLM on time series data. The total objective function is defined as:
\begin{equation}
	\mathcal{L}_{\text{OBJ}}=\mathcal{L}_{\text{TIME}} + \lambda\mathcal{L}_{\text{FEAT}},
	\label{eq_total}
\end{equation}
where
\begin{equation}
	\left\{
	\begin{aligned}
		&\mathcal{L}_{\text{TIME}}=\frac{1}{T}\sum_{i=1}^{T}\|\mathbf{Y}_{i}-\hat{\mathbf{Y}}_{i}\|_2, \\
		&\mathcal{L}_{\text{FEAT}}=\frac{1}{C}\sum_{j=1}^{C}\|\mathbf{F}^{j}_{\text{MS}}-\mathbf{F}^{j}_{\text{T2T}}\|_2.
	\end{aligned}
	\right.
\end{equation}
Here, $\lambda>0$ is a trade-off parameter to balance the contributation of $\mathcal{L}_{\text{TIME}}$ and $\mathcal{L}_{\text{FEAT}}$, $\mathcal{L}_{\text{TIME}}$ encourages the LLM to forecast time series $\hat{\mathbf{Y}}$ that closely matching the ground truth $\mathbf{Y}$, $\mathcal{L}_{\text{FEAT}}$ enriches the multi-scale features $\mathbf{F}_{\text{MS}}$ through semantic alignment with the T2T-generated features $\mathbf{F}_{\text{T2T}}$.

\begin{table*}[ht]
\caption{Average results of multivariate long-term forecasting across four different prediction lengths $T$ in the set $\{96, 192, 336, 720\}$, and the full results are shown in Tab.~\ref{tab::long-term_appendix}. The best and second best results are in \best{bold} and \second{underlined}, respectively. The row ``$1^{\text{st}}$ \textit{Count}" records the times of each method achieving the top results. Here, we reproduce the methods with superscript ``$^\star$" according to their official codebase with the identical experimental setups as ours for fairness comparison. Results for other compared methods are from~\cite{liu2023itransformer}.}
\label{tab::long-term}
\begin{threeparttable}
	\resizebox{\linewidth}{!}{
	\begin{tabular}{c|cc|cc|cc|cc|cc|cc|cc|cc|cc|cc|cc|cc}
		\toprule
		\multicolumn{1}{c}{\multirow{2}{*}{\scalebox{1.1}{Models}}} & \multicolumn{2}{c}{LLM-PS} & \multicolumn{2}{c}{CALF$^\star$} & \multicolumn{2}{c}{TimeLLM$^\star$} & \multicolumn{2}{c}{GPT4TS$^\star$} & \multicolumn{2}{c}{PatchTST} & \multicolumn{2}{c}{iTransformer} & \multicolumn{2}{c}{Crossformer} & \multicolumn{2}{c}{FEDformer} & \multicolumn{2}{c}{TimesNet} & \multicolumn{2}{c}{MICN} & \multicolumn{2}{c}{DLinear} & \multicolumn{2}{c}{TiDE} \\ 
		 \multicolumn{1}{c}{} & \multicolumn{2}{c}{\scalebox{0.8}{\textbf{Ours}}} & \multicolumn{2}{c}{\scalebox{0.8}{\citeyearpar{liu2024taming}}} & \multicolumn{2}{c}{\scalebox{0.8}{\citeyearpar{jintime}}} & \multicolumn{2}{c}{\scalebox{0.8}{\citeyearpar{zhou2023onefitsall}}} & \multicolumn{2}{c}{\scalebox{0.8}{\citeyearpar{nie2022pathtst}}} & \multicolumn{2}{c}{\scalebox{0.8}{\citeyearpar{liu2023itransformer}}} & \multicolumn{2}{c}{\scalebox{0.8}{\citeyearpar{zhang2022crossformer}}} & \multicolumn{2}{c}{\scalebox{0.8}{\citeyearpar{zhou2022fedformer}}} & \multicolumn{2}{c}{\scalebox{0.8}{\citeyearpar{wu2023timesnet}}} & \multicolumn{2}{c}{\scalebox{0.8}{\citeyearpar{wang2022micn}}} & \multicolumn{2}{c}{\scalebox{0.8}{\citeyearpar{zeng2023dlinear}}} & \multicolumn{2}{c}{\scalebox{0.8}{\citeyearpar{das2023tide}}} \\
		\cmidrule(lr){2-3} \cmidrule(lr){4-5} \cmidrule(lr){6-7} \cmidrule(lr){8-9} \cmidrule(lr){10-11} \cmidrule(lr){12-13} \cmidrule(lr){14-15} \cmidrule(lr){16-17} \cmidrule(lr){18-19} \cmidrule(lr){20-21} \cmidrule(lr){22-23} \cmidrule(lr){24-25}
		
		\multicolumn{1}{c}{Metric}& MSE & MAE & MSE & MAE & MSE & MAE & MSE & MAE & MSE & MAE & MSE & MAE & MSE & MAE & MSE & MAE & MSE & MAE & MSE & MAE & MSE & MAE & MSE & MAE \\
		
		\toprule
		
		ETTm1 &\best{0.354} &\best{0.376} & 0.395 & \second{0.390} & 0.410 & 0.409 & 0.389 & 0.397 & \second{0.381} & 0.395 & 0.407 & 0.411 & 0.502 & 0.502 & 0.448 & 0.452 & 0.400 & 0.406 & 0.392 & 0.413 & 0.403 & 0.407 & 0.412 & 0.406 \\
		
		\midrule
		
		ETTm2 &\best{0.262} &\best{0.314} & \second{0.281} & \second{0.321} & 0.296 & 0.340 & 0.285 & 0.331 & 0.285 & 0.327 & 0.291 & 0.335 & 1.216 & 0.707 & 0.305 & 0.349 & 0.291 & 0.333 & 0.328 & 0.382 & 0.350 & 0.401 & 0.289 & 0.326 \\
		
		\midrule
		
		ETTh1 &\best{0.418} &\best{0.420} & \second{0.432} & \second{0.428} & 0.460 & 0.449 & 0.447 & 0.436 & 0.450 & 0.441 & 0.455 & 0.448 & 0.620 & 0.572 & 0.440 & 0.460 & 0.458 & 0.450 & 0.558 & 0.535 & 0.456 & 0.452 & 0.445 & 0.432 \\
		
		\midrule
		ETTh2 &\second{0.350} &\second{0.390} & \best{0.349} & \best{0.382} & 0.389 & 0.408 & 0.381 & 0.408 & 0.366 & 0.394 & 0.381 & 0.405 & 0.942 & 0.684 & 0.437 & 0.449 & 0.414 & 0.427 & 0.587 & 0.525 & 0.559 & 0.515 & 0.611 & 0.550 \\
		
		\midrule
		
		Weather &\best{0.238} &\best{0.269} & 0.250 & \second{0.274} & 0.274 & 0.290 & 0.264 & 0.284 & 0.258 & 0.280 & 0.257 & 0.279 & 0.259 & 0.315 & 0.309 & 0.360 & 0.259 & 0.287 & \second{0.242} & 0.299 & 0.265 & 0.317 & 0.271 & 0.320 \\
		
		\midrule
		
		Electricity &\best{0.161} &\best{0.254} & \second{0.175} & \second{0.265} & 0.223 & 0.309 & 0.205 & 0.290 & 0.216 & 0.304 & 0.178 & 0.270 & 0.244 & 0.334 & 0.214 & 0.327 & 0.192 & 0.295 & 0.186 & 0.294 & 0.212 & 0.300 & 0.251 & 0.344 \\
		
		\midrule

		Traffic &\best{0.427} &\best{0.279} & 0.439 & \second{0.281} & 0.541 & 0.358 & 0.488 & 0.317 & 0.555 & 0.361 & \second{0.428} & 0.282 & 0.550 & 0.304 & 0.610 & 0.376 & 0.620 & 0.336 & 0.541 & 0.315 & 0.625 & 0.383 & 0.760 & 0.473 \\
		\midrule
		
		ILI &\best{1.735} &\second{0.854} & 1.861 & 0.924 & \second{1.829} & 0.924 & 1.871 & \best{0.852} & 2.145 & 0.897 & 2.258 &0.957 & 3.749 & 1.284 & 2.705 & 1.097 & 2.267 & 0.927 & 2.985 & 1.186 & 4.453 & 1.553 & 5.216 & 1.614 \\
		\midrule
		
		ECG &\best{0.225} &\best{0.250} &0.258 &\second{0.260} &0.250 &0.264 &0.262 &\second{0.260} &0.253 &0.277 &0.257 &0.271 &\second{0.244} &0.269 &0.255 &0.279 &0.291 &0.305 &0.305 &0.314 &0.291 &0.307 &0.291 &0.307 \\
		\midrule
		
	\multicolumn{1}{c|}{$1^{\text{st}}$ \emph{Count}} & \multicolumn{2}{c|}{\best{15}}& \multicolumn{2}{c|}{2} & \multicolumn{2}{c|}{0} & \multicolumn{2}{c|}{1} & \multicolumn{2}{c|}{0} & \multicolumn{2}{c|}{0} & \multicolumn{2}{c|}{0} & \multicolumn{2}{c|}{0} & \multicolumn{2}{c|}{0} & \multicolumn{2}{c|}{0} & \multicolumn{2}{c|}{0} & \multicolumn{2}{c}{0} \\
		
		\toprule
		
	\end{tabular}}
\end{threeparttable}
\end{table*}
\begin{table*}[!t]
	\setlength{\tabcolsep}{2pt}
	\tiny
	\centering
	\caption{Results of short-term forecasting across monthly, quarterly, yearly, and others subsets. The input and output time series lengths are $[12, 96]$ and $[6, 48]$, respectively.}
	\begin{threeparttable}
			\resizebox{\linewidth}{!}{
		\begin{tabular}{c|c|cccccccccccccccc}
			\toprule
			
			\multicolumn{2}{c}{\multirow{1}{*}{\scalebox{1.1}{Models}}}&LLM-PS& CALF & TimeLLM & GPT4TS & PatchTST &iTransformer & ETSformer & FEDformer & Autoformer & TimesNet & TCN & N-HiTS & N-BEATS & DLinear & LSSL & LSTM  \\ 
			\multicolumn{2}{c}{} & \scalebox{0.8}{(\textbf{Ours})} &
			\scalebox{0.8}{\citeyearpar{liu2024taming}} &
			\scalebox{0.8}{\citeyearpar{jintime}} & \scalebox{0.8}{\citeyearpar{zhou2023onefitsall}} & \scalebox{0.8}{\citeyearpar{nie2022pathtst}} 	& \scalebox{0.8}{\citeyearpar{liu2023itransformer}}& \scalebox{0.8}{\citeyearpar{woo2022etsformer}} & \scalebox{0.8}{\citeyearpar{zhou2022fedformer}} & \scalebox{0.8}{\citeyearpar{wu2021autoformer}} & \scalebox{0.8}{\citeyearpar{wu2023timesnet}} & \scalebox{0.8}{\citeyearpar{BaiTCN2018}} & \scalebox{0.8}{\citeyearpar{challu2022nhits}} & \scalebox{0.8}{\citeyearpar{oreshkin2019nbeats}} & \scalebox{0.8}{\citeyearpar{zeng2023dlinear}} 
			& \scalebox{0.8}{\citeyearpar{gu2022lssl}} & \scalebox{0.8}{\citeyearpar{Hochreiter1997lstm}}
		
			\\
			\toprule
			
			\multirow{3}{*}{\rotatebox[origin=c]{90}{\scalebox{0.7}{Yearly}}} 
			
			& \scalebox{0.8}{SMAPE} &\best{13.277} & \second{13.314} & 13.419 & 13.531 & 13.477 & 14.252 & 18.009 & 13.728 & 13.974 & 13.387 & 14.920 & 13.418 & 13.436 & 16.965 & 16.675 & 176.040 \\
			
			& \scalebox{0.8}{MASE} &\best{2.973} & 3.009  & 3.005 & 3.015  & 3.019 & 3.208  & 4.487  & 3.048 & 3.134 & \second{2.996}  & 3.364 & 3.045  & 3.043  & 4.283 & 19.953 & 31.033 \\
			
			& \scalebox{0.8}{OWA} &\best{0.780}  & \second{0.786}  & 0.789 & 0.793  & 0.792 & 0.840  & 1.115  & 0.803 & 0.822 & \second{0.786}  & 0.880 & 0.793  & 0.794  & 1.058 & 4.397 & 9.290\\
			
			\midrule
			
			\multirow{3}{*}{\rotatebox[origin=c]{90}{\scalebox{0.7}{Quarterly}}} 
			
			& \scalebox{0.8}{SMAPE} &\best{9.995}& \second{10.049}  & 10.110 & 10.177 & 10.380 & 10.755 & 13.376 & 10.792 & 11.338 & 10.100 & 11.122 & 10.202 & 10.124 & 12.145 & 65.999 & 172.808  \\
			
			& \scalebox{0.8}{MASE} &\best{1.164} & \second{1.166}  & 1.178 & 1.194  & 1.233 & 1.284 & 1.906  & 1.283  & 1.365  & 1.182  & 1.360  & 1.194  & 1.169  & 1.520 & 17.662 & 19.753   \\
			
			& \scalebox{0.8}{OWA} &\second{0.878}  & \best{0.871}  & 0.889 & 0.898  & 0.921 & 0.957 & 1.302  & 0.958  & 1.012  & 0.890  & 1.001  & 0.899  & 0.886  & 1.106 & 9.436 & 15.049  \\
			
			\midrule

			\multirow{3}{*}{\rotatebox[origin=c]{90}{\scalebox{0.7}{Monthly}}}
			
			& \scalebox{0.8}{SMAPE}&\best{12.585} & \second{12.624} & 12.980 & 12.894 & 12.959 & 13.721 & 14.588 & 14.260 & 13.958 & 12.679 & 15.626 & 12.791 & 12.677 & 13.514 & 64.664 & 143.237  \\
			
			& \scalebox{0.8}{MASE} &\best{0.924} & \second{0.922}  & 0.963 & 0.956  & 0.970 & 1.074 & 1.368  & 1.102  & 1.103  & 0.933  & 1.274  & 0.969  & 0.937  & 1.037 & 16.245 & 16.551  \\
			
			& \scalebox{0.8}{OWA} &\best{0.871}  & \best{0.871}  & 0.903 & 0.897  & 0.905 & 0.981 & 1.149  & 1.012  & 1.002  &\second{0.878}  & 1.141  & 0.899  & 0.880  & 0.956 & 9.879 & 12.747   \\
			
			\midrule
			
			\multirow{3}{*}{\rotatebox[origin=c]{90}{\scalebox{0.7}{Others}}}
			
			& \scalebox{0.8}{SMAPE}&\best{4.550} & \second{4.773}  & 4.795 & 4.940  & 4.952 & 5.615 & 7.267  & 4.954  & 5.485  & 4.891  & 7.186  & 5.061  & 4.925  & 6.709 & 121.844 & 186.282 \\
			
			& \scalebox{0.8}{MASE} &\best{3.089} & \second{3.119}  & 3.178 & 3.228  & 3.347 & 3.977 & 5.240  & 3.264  & 3.865  & 3.302  & 4.677  & 3.216  & 3.391  & 4.953 & 91.650 & 119.294  \\
			
			& \scalebox{0.8}{OWA} &\best{0.966}  & \second{0.990}  & 1.006 & 1.029 & 1.049 & 1.218 & 1.591  & 1.036  & 1.187  & 1.035  & 1.494  & 1.040  & 1.053  & 1.487 & 27.273 & 38.411   \\
			
			\midrule
			
			\multirow{3}{*}{\rotatebox[origin=c]{90}{\scalebox{0.7}{Average}}}
			
			& \scalebox{0.8}{SMAPE}&\best{11.721} & \second{11.770} & 11.983 & 11.991 & 12.059 & 12.726 & 14.718 & 12.840 & 12.909 & 11.829 & 13.961 & 11.927 & 11.851 & 13.639 & 67.156 & 160.031  \\
			
			& \scalebox{0.8}{MASE} &\best{1.561} & \second{1.570}  & 1.595 & 1.600  & 1.623 & 15.336 & 2.408  & 1.701  & 1.771  & 1.585  & 1.945  & 1.613  & 1.599  & 2.095 & 21.208 & 25.788   \\
			
			& \scalebox{0.8}{OWA} &\best{0.840}  & \second{0.845}  & 0.859 & 0.861  & 0.869 & 0.929 & 1.172  & 0.918  & 0.939  & 0.851  & 1.023  & 0.861  & 0.855  & 1.051 & 8.021 & 12.642   \\
			
			\midrule
			
			\multicolumn{2}{c}{$1^{\text{st}}$ \emph{Count}} & \best{14} & 2 & 0 & 0 & 0 & 0 & 0 & 0 & 0 & 0 & 0 & 0 & 0 & 0 & 0 & 0 \\
			
			\toprule
			
		\end{tabular}}

	\end{threeparttable}
	\label{tab::short-term_appendix}
\end{table*}
\section{Experiments}
\label{sec_exp}
To demonstrate the effectiveness of our proposed LLM-PS, we conduct intensive experiments on multiple widely used time series datasets for various tasks, including long-term, short-term, few-shot, and zero-shot forecasting.\\
\textbf{Baselines.} We compare our LLM-PS with a large range of SOTA methods, mainly including: 1) LLM-based methods: CALF~\citeyearpar{liu2024taming}, GPT4TS~\citeyearpar{zhou2023onefitsall}, and TimeLLM~\citeyearpar{jintime}; 2) Transformer-based methods: Crossformer~\citeyearpar{zhang2022crossformer}, FEDformer~\citeyearpar{zhou2022fedformer}, PatchTST~\citeyearpar{nie2022pathtst}, iTransformer~\citeyearpar{liu2023itransformer}, ETSformer~\citeyearpar{woo2022etsformer}, and Autoformer~\citeyearpar{wu2021autoformer}; 3) CNN-based methods: TimesNet~\citeyearpar{wu2023timesnet}, TCN~\citeyearpar{BaiTCN2018}, and MICN~\citeyearpar{wang2022micn}; 4) MLP-based methods: DLinear~\citeyearpar{zeng2023dlinear}, TiDE~\citeyearpar{das2023tide}, and TimeMixer~\citeyearpar{wangtimemixer}. Besides, classical works N-HiTS \citeyearpar{challu2022nhits} and N-BEATS \citeyearpar{oreshkin2019nbeats} are also included for short-term forecasting. The details of these comparison methods can be found in the \textbf{Appendix}~\ref{sec_apx_btsfm}.\\
\textbf{Implementation Details.} We follow~\cite{zhou2023onefitsall,liu2024taming} and employ the pre-trained GPT2 model~\cite{radford2019language} (the first six layers) as the default LLM backbone. Our LLM-PS utilizes Adam as the optimizier and the learning rate is 0.0005. For the parameter-efficient LoRA, its rank, scale factor, and dropout ratio are set to 8, 32, and 0.1, respectively. Additionally, the trade-off parameter $\lambda$ in Eq.~\eqref{eq_total} is set to 0.01 and its parametric sensitivities are analyzed in \textbf{Appendix}~\ref{sec_apx_ps}. 
\begin{table*}[!ht]
	\setlength{\tabcolsep}{2.3pt}
	\tiny
	\centering
	\caption{Average results of few-shot forecasting with 10\% training data, where the prediction lengths $T \in \{96, 192, 336, 720\}$ and the full results are provided in Tab.~\ref{tab::few-short}.}
	\label{tab:few-shot}
	\begin{threeparttable}
		\resizebox{\linewidth}{!}{
		\begin{tabular}{c|c|cc|cc|cc|cc|cc|cc|cc|cc|cc|cc|cc}
			\toprule
			
			\multicolumn{2}{c}{\multirow{2}{*}{\scalebox{1.1}{Models}}} & \multicolumn{2}{c}{LLM-PS} & \multicolumn{2}{c}{CALF} & \multicolumn{2}{c}{TimeLLM} & \multicolumn{2}{c}{GPT4TS} & \multicolumn{2}{c}{PatchTST} & \multicolumn{2}{c}{Crossformer} & \multicolumn{2}{c}{FEDformer} & \multicolumn{2}{c}{TimesNet} & \multicolumn{2}{c}{MICN} & \multicolumn{2}{c}{DLinear} & \multicolumn{2}{c}{TiDE}  \\ 
			\multicolumn{2}{c}{} & \multicolumn{2}{c}{\scalebox{0.8}{\textbf{Ours}}}& \multicolumn{2}{c}{\scalebox{0.8}{\citeyearpar{liu2024taming}}} & \multicolumn{2}{c}{\scalebox{0.8}{\citeyearpar{jintime}}} & \multicolumn{2}{c}{\scalebox{0.8}{\citeyearpar{zhou2023onefitsall}}} & \multicolumn{2}{c}{\scalebox{0.8}{\citeyearpar{nie2022pathtst}}} & \multicolumn{2}{c}{\scalebox{0.8}{\citeyearpar{zhang2022crossformer}}} & \multicolumn{2}{c}{\scalebox{0.8}{\citeyearpar{zhou2022fedformer}}} & \multicolumn{2}{c}{\scalebox{0.8}{\citeyearpar{wu2023timesnet}}} & \multicolumn{2}{c}{\scalebox{0.8}{\citeyearpar{wang2022micn}}} & \multicolumn{2}{c}{\scalebox{0.8}{\citeyearpar{zeng2023dlinear}}} & \multicolumn{2}{c}{\scalebox{0.8}{\citeyearpar{das2023tide}}} \\
			
			\cmidrule(lr){3-4} \cmidrule(lr){5-6} \cmidrule(lr){7-8} \cmidrule(lr){9-10} \cmidrule(lr){11-12} \cmidrule(lr){13-14} \cmidrule(lr){15-16} \cmidrule(lr){17-18} \cmidrule(lr){19-20} \cmidrule(lr){21-22} \cmidrule(lr){23-24}
			
			\multicolumn{2}{c}{Metric}& MSE & MAE & MSE & MAE & MSE & MAE & MSE & MAE & MSE & MAE & MSE & MAE & MSE & MAE & MSE & MAE & MSE & MAE & MSE & MAE & MSE & MAE \\
			
			\toprule
			
			\multicolumn{2}{c|}{\scalebox{0.9}{ETTm1}} &\best{0.497} &\best{0.454} & \second{0.504} & \second{0.462} & 0.636 & 0.512 & 0.608 & 0.500 & 0.557 & 0.483 & 1.340 & 0.848 & 0.696 & 0.572 & 0.673 & 0.534 & 0.970 & 0.674 & 0.567 & 0.499 & 0.515 &0.469 \\
			
			\midrule
			
			\multicolumn{2}{c|}{\scalebox{0.9}{ETTm2}} &\best{0.281} &\best{0.324} & 0.302 & \second{0.330} & 0.348 & 0.343 & 0.303 & 0.336 & \second{0.295} & 0.334 & 1.985 & 1.048 & 0.356 & 0.392 & 0.321 & 0.354 & 1.073 & 0.716 & 0.329 & 0.382 & 0.303 & 0.337 \\
			
			\midrule
			
			\multicolumn{2}{c|}{\scalebox{0.9}{ETTh1}} &\best{0.632} &\second{0.546} & \second{0.644} & \best{0.541} & 0.765 & 0.584 & 0.689 & 0.555 & 0.683 & 0.546 & 1.744 & 0.914 & 0.750 & 0.607 & 0.865 & 0.625 & 1.405 & 0.814 & 0.647 & 0.552 & 0.779 & 0.604 \\
			
			\midrule
			\multicolumn{2}{c|}{\scalebox{0.9}{ETTh2}} &\best{0.409} &\best{0.420} & \second{0.419} & \second{0.427} & 0.589 & 0.498 & 0.579 & 0.497 & 0.550 & 0.487 & 3.139 & 1.378 & 0.553 & 0.525 & 0.476 & 0.463 & 2.533 & 1.158 & 0.441 & 0.458 & 0.421 & 0.428 \\
			
			\midrule
			
			\multicolumn{2}{c|}{$1^{\text{st}}$ \emph{Count}} & \multicolumn{2}{c|}{\best{7}}& \multicolumn{2}{c|}{\second{1}} & \multicolumn{2}{c|}{0} & \multicolumn{2}{c|}{0} & \multicolumn{2}{c|}{0} & \multicolumn{2}{c|}{0} & \multicolumn{2}{c|}{0} & \multicolumn{2}{c|}{0} & \multicolumn{2}{c|}{0} & \multicolumn{2}{c|}{0} & \multicolumn{2}{c}{0} \\
			\bottomrule
			
		\end{tabular}}
	\end{threeparttable}
\end{table*}
\begin{table*}[!ht]
	\setlength{\tabcolsep}{2.3pt}
	\tiny
	\centering
	\caption{Average results of zero-shot forecasting, where prediction lengths $T \in \{96, 192, 336, 720\}$ and Tab.~\ref{tab::zero-short} shows the full results. The term ``ETTh1 $\to$ ETTm1" indicates that models trained on the ETTh1 dataset and are evaluated on the ETTm1 dataset, the convention is also followed for other terms.}
	\label{tab:zero-shot}
	\begin{threeparttable}
		\resizebox{\linewidth}{!}{
		\begin{tabular}{c|c|cc|cc|cc|cc|cc|cc|cc|cc|cc|cc|cc}
			\toprule
			
			\multicolumn{2}{c}{\multirow{2}{*}{\scalebox{1.1}{Models}}}& \multicolumn{2}{c}{LLM-PS} & \multicolumn{2}{c}{CALF} & \multicolumn{2}{c}{TimeLLM} & \multicolumn{2}{c}{GPT4TS} & \multicolumn{2}{c}{PatchTST} & \multicolumn{2}{c}{Crossformer} & \multicolumn{2}{c}{FEDformer} & \multicolumn{2}{c}{TimesNet} & \multicolumn{2}{c}{MICN} & \multicolumn{2}{c}{DLinear} & \multicolumn{2}{c}{TiDE}  \\ 
			\multicolumn{2}{c}{} & \multicolumn{2}{c}{\scalebox{0.8}{\textbf{Ours}}}& \multicolumn{2}{c}{\scalebox{0.8}{\citeyearpar{liu2024taming}}} & \multicolumn{2}{c}{\scalebox{0.8}{\citeyearpar{jintime}}} & \multicolumn{2}{c}{\scalebox{0.8}{\citeyearpar{zhou2023onefitsall}}} & \multicolumn{2}{c}{\scalebox{0.8}{\citeyearpar{nie2022pathtst}}} & \multicolumn{2}{c}{\scalebox{0.8}{\citeyearpar{zhang2022crossformer}}} & \multicolumn{2}{c}{\scalebox{0.8}{\citeyearpar{zhou2022fedformer}}} & \multicolumn{2}{c}{\scalebox{0.8}{\citeyearpar{wu2023timesnet}}} & \multicolumn{2}{c}{\scalebox{0.8}{\citeyearpar{wang2022micn}}} & \multicolumn{2}{c}{\scalebox{0.8}{\citeyearpar{zeng2023dlinear}}} & \multicolumn{2}{c}{\scalebox{0.8}{\citeyearpar{das2023tide}}} \\
			\cmidrule(lr){3-4} \cmidrule(lr){5-6} \cmidrule(lr){7-8} \cmidrule(lr){9-10} \cmidrule(lr){11-12} \cmidrule(lr){13-14} \cmidrule(lr){15-16} \cmidrule(lr){17-18} \cmidrule(lr){19-20} \cmidrule(lr){21-22} \cmidrule(lr){23-24}
			
			\multicolumn{2}{c}{Metric} & MSE & MAE& MSE & MAE & MSE & MAE & MSE & MAE & MSE & MAE & MSE & MAE & MSE & MAE & MSE & MAE & MSE & MAE & MSE & MAE & MSE & MAE \\
			
			\toprule
			
			\multicolumn{2}{c|}{\scalebox{0.9}{ETTh1 $\rightarrow$ ETTm1}} &\best{0.721} &\best{0.541} & \second{0.755} & \second{0.574} & 0.847 & 0.565 & 0.798 & \second{0.574} & 0.894 & 0.610 & 0.999 & 0.736 & 0.765 & 0.588 & 0.794 & 0.575 & 1.439 & 0.780 & 0.760 & 0.577 & 0.774 & 0.574 \\
			
			\midrule
			
			\multicolumn{2}{c|}{\scalebox{0.9}{ETTh1 $\rightarrow$ ETTm2}} &\second{0.316} &0.361 & \second{0.316} & \best{0.355} & 0.315 & 0.357 & 0.317 & \second{0.359} & 0.318 & 0.362 & 1.120 & 0.789 & 0.357 & 0.403 & 0.339 & 0.370 & 2.428 & 1.236 & 0.399 & 0.439 & \best{0.314} & \best{0.355} \\
			
			\midrule
			
			\multicolumn{2}{c|}{\scalebox{0.9}{ETTh2 $\rightarrow$ ETTm1}}&\best{0.714} &\best{0.552}  & 0.836 & \second{0.586} & 0.868 & 0.595 & 0.920 & 0.610 & 0.871 & 0.596 & 1.195 & 0.711 & \second{0.741} & 0.588 & 1.286 & 0.705 & 0.764 & 0.601 & 0.778 & 0.594 & 0.841 & 0.590 \\
			
			\midrule
			\multicolumn{2}{c|}{\scalebox{0.9}{ETTh2 $\rightarrow$ ETTm2}} &0.322 &\best{0.359} & \best{0.319} & \second{0.360} & 0.322 & 0.363 & 0.331 & 0.371 & 0.420 & 0.433 & 2.043 & 1.124 & 0.365 & 0.405 & 0.361 & 0.390 & 0.527 & 0.519 & 0.496 & 0.496 & \second{0.321} & 0.364 \\
			
			\midrule
			
				\multicolumn{2}{c|}{$1^{\text{st}}$ \emph{Count}} & \multicolumn{2}{c|}{\best{5}}& \multicolumn{2}{c|}{\second{2}} & \multicolumn{2}{c|}{0} & \multicolumn{2}{c|}{0} & \multicolumn{2}{c|}{0} & \multicolumn{2}{c|}{0} & \multicolumn{2}{c|}{0} & \multicolumn{2}{c|}{0} & \multicolumn{2}{c|}{0} & \multicolumn{2}{c|}{0} & \multicolumn{2}{c}{\second{2}} \\
				
			\bottomrule
			
		\end{tabular}}
	\end{threeparttable}
\end{table*}
\subsection{Long-Term Forecasting}
\textbf{Setups.} We conduct intensive experiments across various popular real-world datasets, including four Electricity Transformer Temperature (ETT) subsets (ETTh1, ETTh2, ETTm1, and ETTm2)~\cite{haoyietalinformer}, Weather, Electricity, Traffic, Illness~\cite{wu2021autoformer}, and Electrocardiography (ECG)~\cite{moody2001impact} datasets. The input length $H$ of time series data is set to 96, with prediction lengths $T$ spanning $\{96, 192, 336, 720\}$. The evaluation metrics are Mean Squared Error (MSE) and Mean Absolute Error (MAE), where lower values of these metrics indicate better model performance. The dataset descriptions are provided in \textbf{Appendix}~\ref{sec_apx_bd}.\\
\textbf{Results.} Tab.~\ref{tab::long-term} reports the brief average results of long-term forecasting with various prediction lengths. Firstly, we can observe that our LLM-PS surpasses the baseline methods in most instances, achieving the top results in 15 of 18 cases. Secondly, compared with the SOTA LLM-based methods, \textit{i.e.}, CALF, TimeLLM, and GPT4TS, our approach achieves consistent MSE/MAE reductions of 6\%/3\%, 11\%/9\%, and 9\%/5\%, respectively. Thirdly, our method significantly outperforms traditional deep learning methods based on Transformer, CNN, and MLP, especially on the Traffic, ILI, and ECG datasets. These results indicate that our LLM-PS can precisely predict long-term time series by effectively leveraging temporal patterns and semantics within the input series with limited length.
\subsection{Short-Term Forecasting}
\textbf{Setups.} We evaluate the M4 dataset~\cite{makridakis2018m4}, which records marketing data in monthly, quarterly, yearly, and others. For short-term forecasting, the prediction horizons are relatively small and range in $[6, 48]$, while the input lengths are twice their corresponding prediction horizons. The evaluation metrics are Mean Absolute Scaled Error (MSAE), Symmetric Mean Absolute Percentage Error (SMAPE), and Overall Weighted Average (OWA). Formal definitions of these metrics are provided in \textbf{Appendix}~\ref{sec_apx_metrics}.\\
\textbf{Results.} The results of short-term forecasting across monthly, quarterly, yearly, and others subsets are shown in Tab.~\ref{tab::short-term_appendix}. Our proposed LLM-PS achieves a SMAPE of 11.721, a MASE of 1.561, and an OWA of 0.840, which consistently outperforms other compared methods. 
\subsection{Few/Zero-Shot Forecasting}
\textbf{Setups.} LLMs possess powerful few-shot and zero-shot learning capabilities~\cite{zhou2023onefitsall,jintime}, which are crucial for time series forecasting in real-world scenarios. We also verify the few-shot and zero-shot learning performance of our LLM-PS. For few-shot forecasting, there are only 10\% training data of ETT datasets are available. For zero-shot forecasting, LLMs trained on one dataset are directly tested on other datasets. The training setups are the same as those in long-term forecasting.
\\
\textbf{Results.} The average results over various prediction horizons for the challenging few-shot forecasting tasks are reported in Tab.~\ref{tab:few-shot}. In such cases, our LLM-PS always achieves the best results in 7 of 8 cases, which demonstrates that our LLM-PS can effectively learn valuable temporal information only using limited data. Compared with the LLM-based methods, \textit{i.e.}, CALF, TimeLLM, and GPT4TS, our method yields 3\%, 17\%, and 13\% performance improvements, respectively.

In addition to the few-shot learning tasks, we also considered the more challenging zero-shot learning tasks. The experimental results are provided in Tab.~\ref{tab:zero-shot}. Our method significantly outperforms the comparison methods in 5 of 8 cases. Notably, our LLM-PS outperforms CALF, TimeLLM, and GPT4TS, which also fine-tune the LLM for time series forecasting, showing performance improvements of 5\%, 8\%, and 9\%, respectively.
\begin{figure*}[ht]
	\centering
	\begin{subfigure}{0.245\textwidth}
		\centering
		\includegraphics[width=\linewidth]{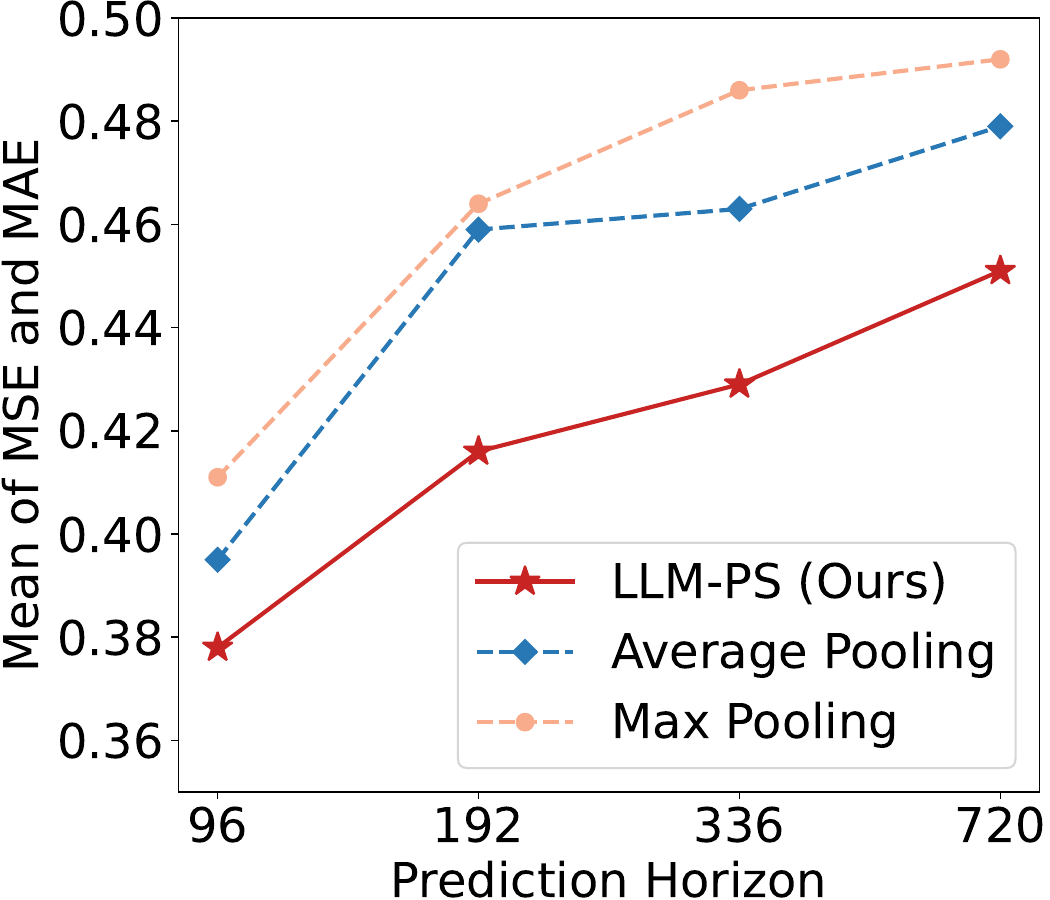}
		\caption{}
		\label{fig_ab_a}
	\end{subfigure}\hfill
	\begin{subfigure}{0.245\textwidth}
		\centering
		\includegraphics[width=\linewidth]{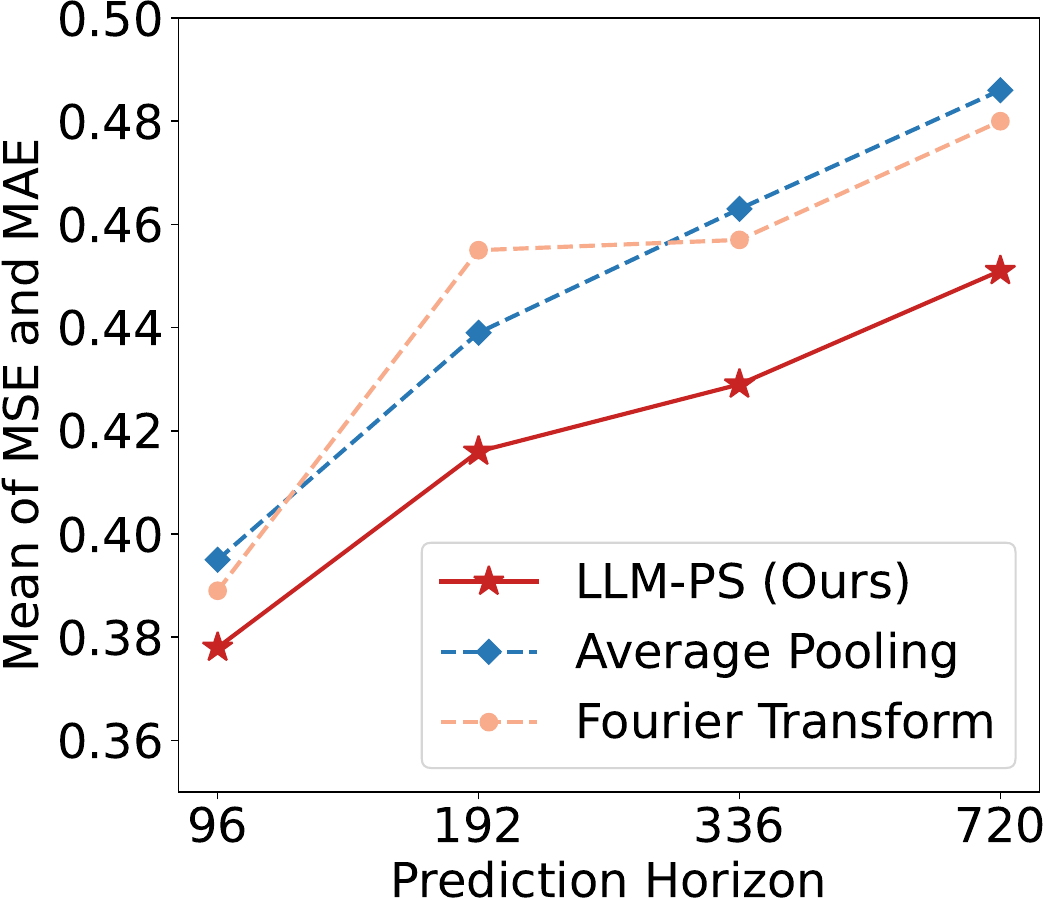}
		\caption{}
		\label{fig_ab_b}
	\end{subfigure}
	\begin{subfigure}{0.245\textwidth}
		\centering
		\includegraphics[width=\linewidth]{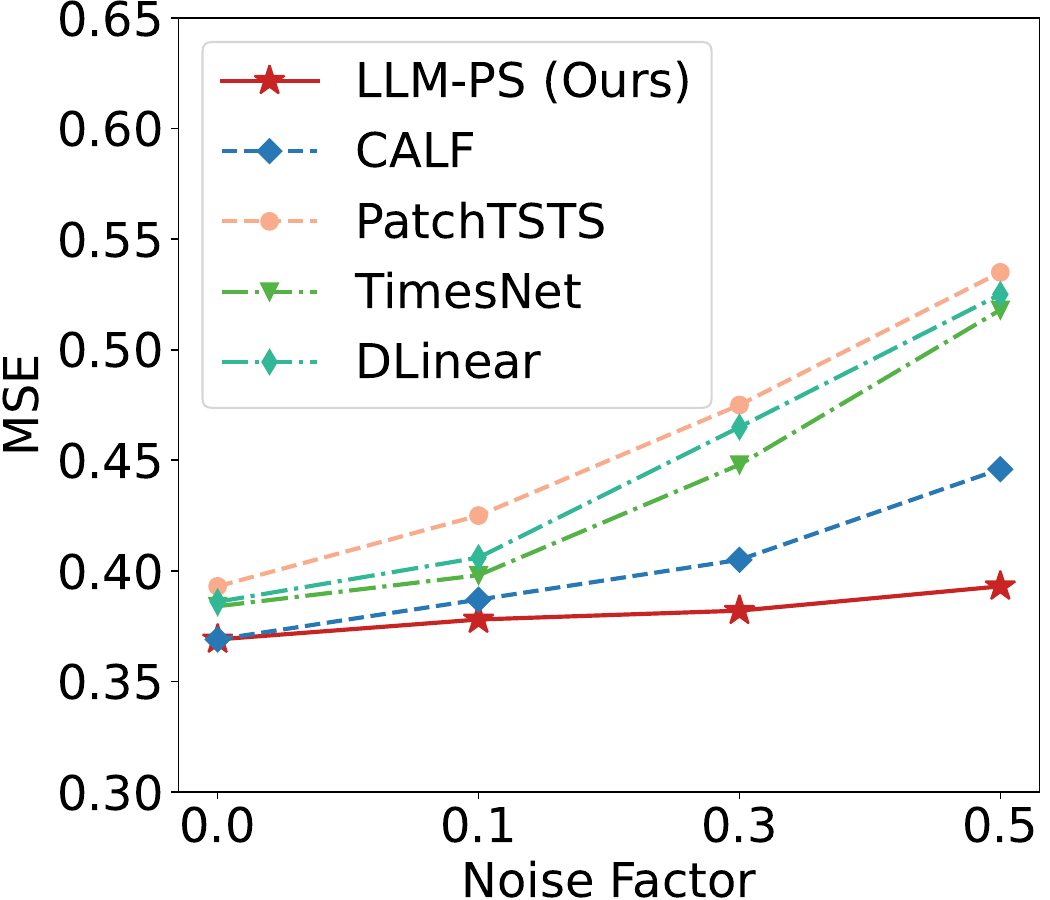}
		\caption{}
		\label{fig_noise_mse}
	\end{subfigure}
	\begin{subfigure}{0.245\textwidth}
		\centering
		\includegraphics[width=\linewidth]{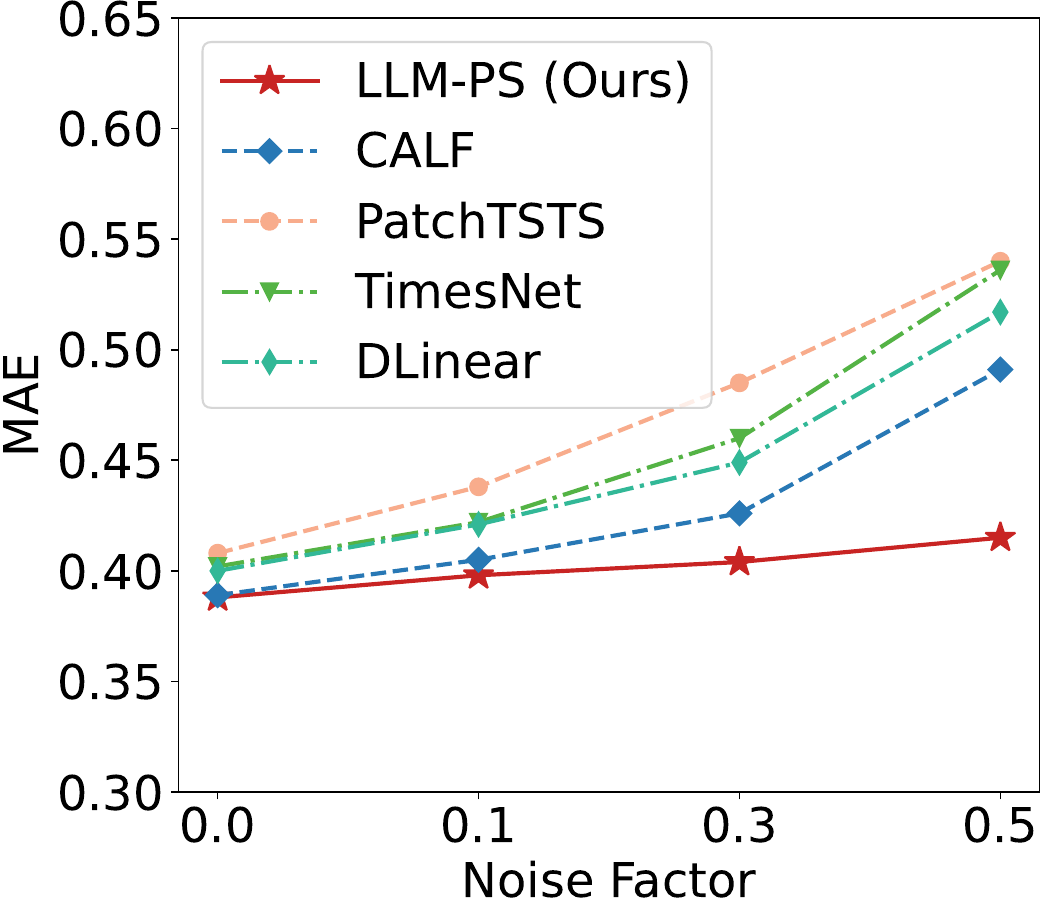}
		\caption{}
		\label{fig_noise_mae}
	\end{subfigure}
	\caption{Analysis of (a) multi-scale feature extraction and (b) temporal patterns decoupling. Subfigures (c) and (d) show the MSE/MAE of various methods on noisy ETTh1 datasets. Notably, lower MSE/MAE indicates better model performance.}
	\label{fig_ab}
\end{figure*}
\subsection{Model Analysis}
\label{sec_exp_ma}
\textbf{Multi-Scale Feature Extraction.} We compare our MSCNN against existing multi-scale features extraction techniques~\cite{kowsher2024llm}, which convert the input time series into diverse scales using average or max pooling layers with varying window sizes. As shown in Fig.~\ref{fig_ab_a}, our method consistently outperforms those employing pooling operations. These results indicate that our method excels in extracting the multi-scale features, thereby effectively capturing the short-term and long-term patterns in input time series. Further analysis is provided in \textbf{Appendix}~\ref{sec_apx_visdtp}.\\
\textbf{Temporal Patterns Decoupling.} We compare our wavelet-transform-based decoupling with the Fourier-transform-based and pooling-based decoupling techniques used in existing methods~\cite{wu2021autoformer,wangtimemixer}. As shown in Fig.~\ref{fig_ab_b}, our decoupling operation based on wavelet transform achieves better performance than those based on Fourier transform and average pooling. The short-term and long-term components decomposed by our LLM-PS and compared methods are visualized in Fig.~\ref{fig_visdtp}. \\
\textbf{Semantic Information Utilization.} Tab.~\ref{tab:ab} reports the results of ablating our proposed T2T module for semantic information learning. It can be found that the performance of LLM decreases in 9 of 10 cases, which demonstrates that the semantic information extracted by our T2T is helpful in enhancing the TSF performance of LLM. The detailed structure of the T2T module is provided in \textbf{Appendix}~\ref{sec_apx_t2t}.
\begin{table}[!h]
	\setlength{\tabcolsep}{5pt}
	\centering
	\tiny
	\caption{Long-term forecasting results of ablating our proposed T2T module on the ETTh1 dataset.}
	\label{tab:ab}
	\begin{threeparttable}
		\resizebox{\linewidth}{!}{
			\begin{tabular}{c|ccccc}
				\toprule
				\multirow{2}{*}{Type} & \multicolumn{5}{c}{MSE / MAE} \\
				& 96 & 192 & 336 & 720 &Mean \\
				\midrule
				w/o T2T      & 0.373 / 0.395 & \best{0.416} / 0.425 & 0.439 / 0.432 &0.464 /0.470  &0.426 /0.431 \\
				LLM-PS   & \best{0.369} / \best{0.388} & 0.418 / \best{0.415} & \best{0.432} / \best{0.426} & \best{0.452} / \best{0.451} & \best{0.418} / \best{0.420}\\
				\toprule
		\end{tabular}}
	\end{threeparttable}
\end{table}\\
\textbf{Model Efficiency.} We evaluate the efficiency of our method with other LLM fine-tuning methods on four datasets, including ETTh1, ETTm1, Weather, and Traffic. For a fair comparison, all experiments adopt the same experimental settings and LLM backbone (\textit{i.e.}, GPT2), where input length, prediction length, and patience for early stopping are set to 96, 96, and 5, respectively. All experiments are conducted on a single NVIDIA RTX 4090 GPU. As reported in Tab.~\ref{tab:time}, our proposed LLM-PS achieves the best performance with significantly lower training costs. Compared with LLMMixer, which also learns multi-scale features from the multiple input series with various scales, our method attains a 17\% performance edge while utilizing just 9\% of the training time. These experimental results indicate that our method can efficiently and effectively learn the temporal patterns and semantic information from time series data, and thus achieving reliable TSF performance.
\begin{table}[!t]
	\centering
	\caption{The fine-tuning costs, mean MSE/MAE across four datasets of our LLM-PS and other LLM-based methods.}
	\label{tab:time}
	\begin{threeparttable}
		\resizebox{\linewidth}{!}{
			\begin{tabular}{c|cccc|cc}
				\toprule
			\multirow{2}{*}{Model}  	& \multicolumn{4}{c|}{Time (s)}  & \multirow{2}{*}{Mean MSE}                                     & \multirow{2}{*}{Mean MAE}        \\
				& ETTh1 & ETTm1 & Weather & Traffic &  &  \\
				\midrule
				GPT4TS~\citeyearpar{zhou2023onefitsall}      & 421  & 1140   & 4565 & 59164  & 0.339 & 0.323  \\
				Time-LLM~\citeyearpar{jintime}   & 2780 & 11929  & 36188 & 465136 & 0.372 & 0.346  \\
				LLMMixer~\citeyearpar{kowsher2024llm}	& 635 & 2493  & 9640 & 10464 & 0.372 & 0.346 \\
				CALF~\citeyearpar{liu2024taming} & 354  & 1394   & 1259   & 4929  & 0.315 & 0.302 \\
				LLM-PS (Ours) & \best{192}  & \best{481}  & \best{260}   & \best{1092}  & \best{0.301} & \best{0.298} \\
				\toprule
		\end{tabular}}
	\end{threeparttable}
	\vspace{-0.3cm}
\end{table}\\
\textbf{Noisy Data.} Time series data in real-world applications is usually noisy due to measurement errors and missing values. In such cases, the target time series is more challenging than that in training data, and the models are hard to predict reliably. To assess the robustness of our LLM-PS against noise, we evaluate it on the ETTh1 dataset with Gaussian noise, where the noise factors are in $[0.0, 0.1, 0.3, 0.5]$. Both the input and prediction lengths are set to 96. As reported in Fig.~\ref{fig_noise_mse}\&\ref{fig_noise_mae}, our method consistently achieves superior performance across various noise factors. In particular, our method outperforms other comparison approaches by an even more significant margin as the noise factor increases. These experimental results demonstrate that our LLM-PS is robust to noise and can be effectively applied to real-world scenarios for time series forecasting.

\section{Conclusion}
\label{sec_clu}
In this paper, we identify the intrinsic characteristics of time series data, \textit{i.e.}, diverse temporal patterns and semantic sparsity. These properties are critical for reliable time series forecasting but are usually neglected by existing LLM-based methods, and thus resulting in suboptimal performance. To address this problem, we propose LLM-PS, a novel TSF framework that learns fundamental temporal patterns and valuable semantics from time series data through the novel MSCNN and T2T modules. As a result, our LLM-PS can comprehensively understand time series data, thereby enabling accurate generation of time series. Our intensive experiments demonstrate that LLM-PS achieves SOTA performance across multiple benchmark datasets spanning critical real-world domains, mainly including finance, energy, transportation, and healthcare.
\section*{Impact Statement}
This paper proposes a novel framework, LLM-PS, which aims to advance time series forecasting using large language models. Our work focuses on integrating temporal patterns and semantics in time series data, enabling state-of-the-art performance in various forecasting scenarios.

We anticipate several potential societal benefits, including improved decision-making in critical areas such as finance, healthcare, and energy management. However, as with any machine learning advancement, ethical considerations must be acknowledged, including the potential misuse of forecasting technologies in domains like financial speculation or surveillance. To mitigate these risks, we encourage responsible and transparent applications of our methods. 

In conclusion, we do not identify any immediate ethical concerns or risks unique to our methodology, and we believe it can benefit society.
\nocite{langley00}

\bibliography{example_paper}
\bibliographystyle{icml2025}

\newpage
\appendix
\onecolumn
\clearpage
\appendix
\section{Full Experimental Results}
\begin{table*}[!ht]
	\setlength{\tabcolsep}{1.35pt}
	\tiny
	\centering
		\caption{Full results for long-term forecasting consider prediction horizons $H$ within $\{96, 192, 336, 720\}$. The term ``\textit{Avg.}" represents the average results across the four prediction lengths. The best and second best outcomes are highlighted in \best{bold} and \second{underlined}, respectively. The notation ``$1^{\text{st}}$ \textit{Count}" denotes the frequency of each method achieving the top results.}
	\begin{threeparttable}
		\begin{tabular}{c|c|>{\columncolor{yellow!15}}c>{\columncolor{yellow!15}}c|>{\columncolor{yellow!15}}c>{\columncolor{yellow!15}}c|>{\columncolor{yellow!15}}c>{\columncolor{yellow!15}}c|>{\columncolor{yellow!15}}c>{\columncolor{yellow!15}}c|>{\columncolor{green!15}}c>{\columncolor{green!15}}c|>{\columncolor{green!15}}c>{\columncolor{green!15}}c|>{\columncolor{green!15}}c>{\columncolor{green!15}}c|>{\columncolor{green!15}}c>{\columncolor{green!15}}c|>{\columncolor{green!15}}c>{\columncolor{green!15}}c|>{\columncolor{green!15}}c>{\columncolor{green!15}}c|>{\columncolor{purple!10}}c>{\columncolor{purple!10}}c|>{\columncolor{purple!10}}c>{\columncolor{purple!10}}c|>{\columncolor{blue!15}}c>{\columncolor{blue!15}}c|>{\columncolor{blue!15}}c>{\columncolor{blue!15}}c}

			\toprule
			
			\multicolumn{2}{c}{Categories} & \multicolumn{8}{>{\columncolor{yellow!15}}c}{LLM-based} & \multicolumn{12}{>{\columncolor{green!15}}c}{Transformer-based} & \multicolumn{4}{>{\columncolor{purple!10}}c}{CNN-based} & \multicolumn{4}{>{\columncolor{blue!15}}c}{MLP-based} \\
			
			\toprule
			
			\multicolumn{2}{c}{\multirow{2}{*}{\scalebox{1.1}{Models}}}& \multicolumn{2}{>{\columncolor{yellow!15}}c}{LLM-PS} & \multicolumn{2}{>{\columncolor{yellow!15}}c}{CALF} & \multicolumn{2}{>{\columncolor{yellow!15}}c}{TimeLLM} & \multicolumn{2}{>{\columncolor{yellow!15}}c}{GPT4TS} & \multicolumn{2}{>{\columncolor{green!15}}c}{PatchTST} & \multicolumn{2}{>{\columncolor{green!15}}c}{iTransformer} & \multicolumn{2}{>{\columncolor{green!15}}c}{Crossformer} & \multicolumn{2}{>{\columncolor{green!15}}c}{FEDformer} & \multicolumn{2}{>{\columncolor{green!15}}c}{Autoformer} & \multicolumn{2}{>{\columncolor{green!15}}c}{Informer} & \multicolumn{2}{>{\columncolor{purple!10}}c}{TimesNet} & \multicolumn{2}{>{\columncolor{purple!10}}c}{MICN} & \multicolumn{2}{>{\columncolor{blue!15}}c}{DLinear} & \multicolumn{2}{>{\columncolor{blue!15}}c}{TiDE} \\
	\multicolumn{2}{c}{} & \multicolumn{2}{>{\columncolor{yellow!15}}c}{\scalebox{0.8}{\textbf{Ours}}} & \multicolumn{2}{>{\columncolor{yellow!15}}c}{\scalebox{0.8}{(\citeyearpar{liu2024taming})}} & \multicolumn{2}{>{\columncolor{yellow!15}}c}{\scalebox{0.8}{\citeyearpar{jintime}}} & \multicolumn{2}{>{\columncolor{yellow!15}}c}{\scalebox{0.8}{\citeyearpar{zhou2023onefitsall}}} & \multicolumn{2}{>{\columncolor{green!15}}c}{\scalebox{0.8}{\citeyearpar{nie2022pathtst}}} & \multicolumn{2}{>{\columncolor{green!15}}c}{\scalebox{0.8}{\citeyearpar{liu2023itransformer}}} & \multicolumn{2}{>{\columncolor{green!15}}c}{\scalebox{0.8}{\citeyearpar{zhang2022crossformer}}} & \multicolumn{2}{>{\columncolor{green!15}}c}{\scalebox{0.8}{\citeyearpar{zhou2022fedformer}}} & \multicolumn{2}{>{\columncolor{green!15}}c}{\scalebox{0.8}{\citeyearpar{wu2021autoformer}}} & \multicolumn{2}{>{\columncolor{green!15}}c}{\scalebox{0.8}{\citeyearpar{zhou2021informer}}} & \multicolumn{2}{>{\columncolor{purple!10}}c}{\scalebox{0.8}{\citeyearpar{wu2023timesnet}}} & \multicolumn{2}{>{\columncolor{purple!10}}c}{\scalebox{0.8}{\citeyearpar{wang2022micn}}} & \multicolumn{2}{>{\columncolor{blue!15}}c}{\scalebox{0.8}{\citeyearpar{zeng2023dlinear}}} & \multicolumn{2}{>{\columncolor{blue!15}}c}{\scalebox{0.8}{\citeyearpar{das2023tide}}} \\

			\cmidrule(lr){3-4} \cmidrule(lr){5-6} \cmidrule(lr){7-8} \cmidrule(lr){9-10} \cmidrule(lr){11-12} \cmidrule(lr){13-14} \cmidrule(lr){15-16} \cmidrule(lr){17-18} \cmidrule(lr){19-20} \cmidrule(lr){21-22} \cmidrule(lr){23-24} \cmidrule(lr){25-26} \cmidrule(lr){27-28} \cmidrule(lr){29-30}
			
			\multicolumn{2}{c}{Metric}& MSE & MAE & MSE & MAE & MSE & MAE & MSE & MAE & MSE & MAE & MSE & MAE & MSE & MAE & MSE & MAE & MSE & MAE & MSE & MAE & MSE & MAE & MSE & MAE & MSE & MAE & MSE & MAE \\
			
			\toprule
			
			\multirow{5}{*}{\rotatebox[origin=c]{90}{ETTm1}} 
			
			& 96 &\best{0.288} &\best{0.334}   & \second{0.323} & \second{0.349} & 0.359 & 0.381 & 0.329 & 0.364 & 0.321 & 0.360 & 0.341 & 0.376 & 0.360 & 0.401 & 0.379 & 0.419 & 0.505 & 0.475 & 0.672 & 0.571 & 0.338 & 0.375 & 0.316 & 0.362 & 0.345 & 0.372 & 0.352 & 0.373 \\
			
			& 192 &\best{0.333} &\best{0.361} & 0.374 & 0.375 & 0.383 & 0.393 & 0.368 & 0.382 & \second{0.362} & 0.384  & 0.382 & 0.395 & 0.402 & 0.440 & 0.426 & 0.441 & 0.553 & 0.496 & 0.795 & 0.669 & \second{0.374} & 0.387 & 0.363 & 0.390 & 0.380 & 0.389 & 0.389 & 0.391 \\
			
			& 336&\best{0.367} &\best{0.386}  & 0.409 & \second{0.399} & 0.416 & 0.414 & 0.400 & 0.403 & \second{0.392} & 0.402 & 0.418 & 0.418 & 0.543 & 0.528 & 0.445 & 0.459 & 0.621 & 0.537 & 1.212 & 0.871 & 0.410 & 0.411 & 0.408 & 0.426 & 0.413 & 0.413 & 0.423 & 0.413 \\
			
			& 720&\best{0.429} &\best{0.424}  & 0.477 & 0.438 & 0.483 & 0.449 & 0.460 & 0.439 & \second{0.450} & \second{0.435} & 0.487 & 0.456 & 0.704 & 0.642 & 0.543 & 0.490 & 0.671 & 0.561 & 1.166 & 0.823 & 0.478 & 0.450 & 0.481 & 0.476 & 0.474 & 0.453 & 0.485 & 0.448 \\
			
			\cmidrule(lr){2-30}
			
			& \emph{Avg.} &\best{0.354} &\best{0.376}& 0.395 & \second{0.390} & 0.410 & 0.409 & 0.389 & 0.397 & \second{0.381} & 0.395 & 0.407 & 0.411 & 0.502 & 0.502 & 0.448 & 0.452 & 0.588 & 0.517 & 0.961 & 0.734 & 0.400 & 0.406 & 0.392 & 0.413 & 0.403 & 0.407 & 0.412 & 0.406 \\
			
			\midrule
			
			\multirow{5}{*}{\rotatebox[origin=c]{90}{ETTm2}} 
			
			& 96 &\best{0.170} &\best{0.254}  & \second{0.178} & \second{0.256} & 0.193 & 0.280 & \second{0.178} & 0.263 & \second{0.178} & 0.260 & 0.185 & 0.272 & 0.273 & 0.356 & 0.203 & 0.287 & 0.255 & 0.339 & 0.365 & 0.453 & 0.187 & 0.267 & 0.179 & 0.275 & 0.193 & 0.292 & 0.181 & 0.264 \\
			
			& 192 &\best{0.224} &\best{0.289} & \second{0.242} & \second{0.297} & 0.257 & 0.318 & 0.245 & 0.306 & 0.249 & 0.307 & 0.253
			& 0.313 &  0.426 & 0.487 & 0.269 & 0.328 & 0.249 & 0.309 & 0.281 & 0.340 & 0.533 & 0.563 & 0.307 & 0.376 & 0.284 & 0.362 & 0.246 & 0.304 \\
			
			& 336 &\best{0.280} &\best{0.327} & \second{0.307} & \second{0.339} & 0.317 & 0.353 & 0.309 & 0.347 & 0.313 & 0.346 & 0.315
			& 0.350 & 1.013 & 0.714 & 0.325 & 0.366 & 0.339 & 0.372 & 1.363 & 0.887 & 0.321 & 0.351 & 0.325 & 0.388 & 0.369 & 0.427 & 0.307 & 0.341 \\
			
			& 720&\best{0.374} &\best{0.384}  & \second{0.397} & \second{0.393} & 0.419 & 0.411 & 0.409 & 0.408 & 0.400 & 0.398 & 0.413
			& 0.406 & 3.154 & 1.274 & 0.421 & 0.415 & 0.433 & 0.432 & 3.379 & 1.338 & 0.408 & 0.403 & 0.502 & 0.490 & 0.554 & 0.522 & 0.407 & 0.397 \\
			
			\cmidrule(lr){2-30}
			
			& \emph{Avg.}&\best{0.262}&\best{0.314} & \second{0.281} & \second{0.321} & 0.296 & 0.340 & 0.285 & 0.331 & 0.285 & 0.327 & 0.291 & 0.335 & 1.216 & 0.707 & 0.305 & 0.349 & 0.327 & 0.371 & 1.410 & 0.810 & 0.291 & 0.333 & 0.328 & 0.382 & 0.350 & 0.401 & 0.289 & 0.326 \\
			
			\midrule
			
			\multirow{5}{*}{\rotatebox[origin=c]{90}{ETTh1}} 
			
			& 96 &\best{0.369} &\best{0.388}  & \best{0.369} & \second{0.389} & 0.398 & 0.410 & \second{0.376} & 0.397 & 0.393 & 0.408 & 0.386
			& 0.404 & 0.420 & 0.439 & \second{0.376} & 0.419 & 0.449 & 0.459 & 0.865 & 0.713 & 0.384 & 0.402 & 0.421 & 0.431 & 0.386 & 0.400 & 0.384 & 0.393 \\
			
			& 192 &\best{0.418} &\best{0.415} & 0.427 & 0.423 & 0.451 & 0.440 & 0.438 & 0.426 & 0.445 & 0.434 & 0.441 & 0.436 & 0.540 & 0.519 & \second{0.420} & 0.448 & 0.436 & 0.429 & 0.500 & 0.482 & 1.008 & 0.792 & 0.474 & 0.487 & 0.437 & 0.432 & 0.436 & \second{0.422} \\
			
			& 336 &\best{0.432} &\best{0.426} & \second{0.456} & \second{0.436} & 0.508 & 0.471 & 0.479 & 0.446 & 0.484 & 0.451 & 0.489 & 0.461 & 0.722 & 0.648 & 0.459 & 0.465 & 0.521 & 0.496 & 1.107 & 0.809 & 0.491 & 0.469 & 0.569 & 0.551 & 0.481 & 0.459 & 0.480 & 0.445 \\
			
			& 720 &\best{0.452} &\best{0.451} & \second{0.479} & \second{0.467} & 0.483 & 0.478 & 0.495 & 0.476 & 0.480 & 0.471 & 0.508 & 0.493 & 0.799 & 0.685 & 0.506 & 0.507 & 0.514 & 0.512 & 1.181 & 0.865 & 0.521 & 0.500 & 0.770 & 0.672 & 0.519 & 0.516 & 0.481 & 0.469 \\
			
			\cmidrule(lr){2-30}
			
			& \emph{Avg.} &\best{0.418} &\best{0.420}& \second{0.432} & \second{0.428} & 0.460 & 0.449 & 0.447 & 0.436 & 0.450 & 0.441 & 0.455 & 0.448 & 0.620 & 0.572 & 0.440 & 0.460 & 0.496 & 0.487 & 1.040 & 0.795 & 0.458 & 0.450 & 0.558 & 0.535 & 0.456 & 0.452 & 0.445 & 0.432 \\
			
			\midrule
			
			\multirow{5}{*}{\rotatebox[origin=c]{90}{ETTh2}} 
			
			& 96 &\best{0.279} &\second{0.341}  & \best{0.279} & \best{0.331} & 0.295 & 0.346 & 0.295 & 0.348 & \second{0.294} & 0.343  & 0.300 & 0.349 & 0.745 & 0.584 & 0.358 & 0.397 & 0.346 & 0.388 & 3.755 & 1.525 & 0.340 & 0.374 & 0.299 & 0.364 & 0.333 & 0.387 & 0.400 & 0.440 \\
			
			& 192 &\second{0.356} &\second{0.387} & \best{0.353} & \best{0.380} & 0.386 & 0.399 & 0.386 & 0.404 & 0.377 & 0.393 & 0.379 & 0.398 & 0.877 & 0.656 & 0.429 & 0.439 & 0.456 & 0.452 & 5.602 & 1.931 & 0.402 & 0.414 & 0.441 & 0.454 & 0.477 & 0.476 & 0.528 & 0.509 \\
			
			& 336 &\best{0.350} &\best{0.393} & \second{0.362} & \second{0.394} & 0.447 & 0.443 & 0.421 & 0.435 & 0.381 & 0.409 & 0.418 & 0.429 & 1.043 & 0.731 & 0.496 & 0.487 & 0.482 & 0.486 & 4.721 & 1.835 & 0.452 & 0.452 & 0.654 & 0.567 & 0.594 & 0.541 & 0.643 & 0.571 \\
			
			& 720&0.413 &0.437  & \best{0.404} & \best{0.426} & 0.428 & 0.444 & 0.422 & 0.445 & \second{0.412} & \second{0.433} & 0.428 & 0.445 & 1.104 & 0.763 & 0.463 & 0.474 & 0.515 & 0.511 & 3.647 & 1.625 & 0.462 & 0.468 & 0.956 & 0.716 & 0.831 & 0.657 & 0.874 & 0.679 \\
			
			\cmidrule(lr){2-30}
			
			& \emph{Avg.}&\second{0.350} &\second{0.390} & \best{0.349} & \best{0.382} & 0.389 & 0.408 & 0.381 & 0.408 & 0.366 & 0.394 & 0.381 & 0.405 & 0.942 & 0.684 & 0.437 & 0.449 & 0.450 & 0.459 & 4.431 & 1.729 & 0.414 & 0.427 & 0.587 & 0.525 & 0.559 & 0.515 & 0.611 & 0.550 \\
			
			\midrule
			
			\multirow{5}{*}{\rotatebox[origin=c]{90}{Weather}}
			
			& 96  &\best{0.157} &\second{0.205} & 0.164 & \best{0.204} & 0.195 & 0.233 & 0.182 & 0.223 & 0.177 & 0.218 & 0.174 & 0.214 &  \second{0.158} & 0.230 & 0.217 & 0.296 & 0.266 & 0.336 & 0.300 & 0.384 & 0.172 & 0.220 & 0.161 & 0.229 & 0.196 & 0.255 & 0.202 & 0.261 \\
			
			& 192 &\best{0.202} &\best{0.245} & 0.214 & \second{0.250} & 0.240 & 0.269 & 0.231 & 0.263 & 0.225 & 0.259 & 0.221 & 0.254 & \second{0.206} & 0.277 & 0.276 & 0.336 & 0.307 & 0.367 & 0.598 & 0.544 & 0.219 & 0.261 & 0.220 & 0.281 & 0.237 & 0.296 & 0.242 & 0.298 \\
			
			& 336&\best{0.255} &\best{0.286}  & \second{0.269} & \second{0.291} & 0.293 & 0.306 & 0.283 & 0.300 & 0.278 & 0.297 & 0.278 & 0.296 & 0.272 & 0.335 & 0.339 & 0.380 & 0.359 & 0.395 & 0.578 & 0.523 & 0.280 & 0.306 & 0.278 & 0.331 & 0.283 & 0.335 & 0.287 & 0.335 \\
			
			& 720 &\second{0.336} &\best{0.338} & 0.355 & 0.352 & 0.368 & 0.354 & 0.360 & 0.350 & 0.354 & \second{0.348} & 0.358 & 0.349 & 0.398 & 0.418 & 0.403 & 0.428 & 0.419 & 0.428 & 1.059 & 0.741 & 0.365 & 0.359 & \best{0.311} & 0.356 & 0.345 & 0.381 & 0.351 & 0.386 \\
			
			\cmidrule(lr){2-30}
			
			& \emph{Avg.}&\best{0.238} &\best{0.269} & 0.250 & \second{0.274} & 0.274 & 0.290 & 0.264 & 0.284 & 0.258 & 0.280 & 0.257 & 0.279 & 0.259 & 0.315 & 0.309 & 0.360 & 0.338 & 0.382 & 0.634 & 0.548 & 0.259 & 0.287 & \second{0.242} & 0.299 & 0.265 & 0.317 & 0.271 & 0.320 \\
			
			\midrule
			
			\multirow{5}{*}{\rotatebox[origin=c]{90}{Electricity}}
			
			& 96  &\best{0.131} &\best{0.222}  & \second{0.145} & \second{0.238} & 0.204 & 0.293 & 0.185 & 0.272 & 0.195 & 0.285 & 0.148 & 0.240 & 0.219 & 0.314 & 0.193 & 0.308 & 0.201 & 0.317 & 0.274 & 0.368 & 0.168 & 0.272 & 0.164 & 0.269 & 0.197 & 0.282 & 0.237 & 0.329 \\
			
			& 192 &\best{0.151} &\best{0.240} & \second{0.161} & \second{0.252} & 0.207 & 0.295 & 0.189 & 0.276 & 0.199 & 0.289 & 0.162 & 0.253 & 0.231 & 0.322 & 0.201 & 0.315 & 0.222 & 0.334 & 0.296 & 0.386 & 0.184 & 0.289 & 0.177 & 0.285 & 0.196 & 0.285 & 0.236 & 0.330 \\
			
			& 336 &\best{0.162} &\best{0.256} & \second{0.175} & \second{0.267} & 0.219 & 0.308 & 0.204 & 0.291 & 0.215 & 0.305 & 0.178 & 0.269 & 0.246 & 0.337 & 0.214 & 0.329 & 0.231 & 0.338 & 0.300 & 0.394 & 0.198 & 0.300 & 0.193 & 0.304 & 0.209 & 0.301 & 0.249 & 0.344 \\
			
			& 720 &\second{0.213} &\best{0.297}   & 0.222 & \second{0.303} & 0.263 & 0.341 & 0.245 & 0.324 & 0.256 & 0.337 & 0.225 & 0.317 & 0.280 & 0.363 & 0.246 & 0.355 & 0.254 & 0.361 & 0.373 & 0.439 & 0.220 & 0.320 & \best{0.212} & 0.321 & 0.245 & 0.333 & 0.284 & 0.373 \\
			
			\cmidrule(lr){2-30}
			
			& \emph{Avg.} &\best{0.164} &\best{0.254} & \second{0.175} & \second{0.265} & 0.223 & 0.309 & 0.205 & 0.290 & 0.216 & 0.304 & 0.178 & 0.270 & 0.244 & 0.334 & 0.214 & 0.327 & 0.227 & 0.338 & 0.311 & 0.397 & 0.192 & 0.295 & 0.186 & 0.294 & 0.212 & 0.300 & 0.251 & 0.344 \\
			
			\midrule
			
			\multirow{5}{*}{\rotatebox[origin=c]{90}{Traffic}}
			
			& 96&\best{0.392} &\best{0.267}   & 0.407 & \second{0.268} & 0.536 & 0.359 & 0.468 & 0.307 & 0.544 & 0.359 & \second{0.395} & \second{0.268} &  0.522 & 0.290 & 0.587 & 0.366 & 0.613 & 0.388 & 0.719 & 0.391 & 0.593 & 0.321 & 0.519 & 0.309 & 0.650 & 0.396 & 0.805 & 0.493 \\
			
			& 192 &\best{0.413} &\best{0.265} & 0.430 & 0.278 & 0.530 & 0.354 & 0.476 & 0.311 & 0.540 & 0.354 & \second{0.417} & \second{0.276} & 0.530 & 0.293 & 0.604 & 0.373 & 0.616 & 0.382 & 0.696 & 0.379 & 0.617 & 0.336 & 0.537 & 0.315 & 0.598 & 0.370 & 0.756 & 0.474 \\
			
			& 336&\second{0.440} &0.282  & 0.444 & \best{0.281} & 0.530 & 0.349 & 0.488 & 0.317 & 0.551 & 0.358 & \best{0.433} & \second{0.283} & 0.558 & 0.305 & 0.621 & 0.383 & 0.622 & 0.337 & 0.777 & 0.420 & 0.629 & 0.336 & 0.534 & 0.313 & 0.605 & 0.373 & 0.762 & 0.477 \\
			
			& 720&\best{0.464} &\best{0.300}  & 0.477 & \best{0.300} & 0.569 & 0.371 & 0.521 & 0.333 & 0.586 & 0.375 & \second{0.467} & \second{0.302} & 0.589 & 0.328 & 0.626 & 0.382 & 0.660 & 0.408 & 0.864 & 0.472 & 0.640 & 0.350 & 0.577 & 0.325 & 0.645 & 0.394 & 0.719 & 0.449 \\
			
			\cmidrule(lr){2-30}
			
			& \emph{Avg.}&\best{0.427} &\best{0.279} & 0.439 & \second{0.281} & 0.541 & 0.358 & 0.488 & 0.317 & 0.555 & 0.361 & \second{0.428} & 0.282 & 0.550 & 0.304 & 0.610 & 0.376 & 0.628 & 0.379 & 0.764 & 0.416 & 0.620 & 0.336 & 0.541 & 0.315 & 0.625 & 0.383 & 0.760 & 0.473 \\
			
			\midrule
			
				\multirow{5}{*}{\rotatebox[origin=c]{90}{ILI}}
			
			& 24 &\best{1.630} &\best{0.798}   &1.672 & 0.841 & \second{1.651} & 0.841 & 1.869 & \second{0.823} & 2.221 & 0.883 & 2.321 & 0.937 &  3.449 & 1.238 & 2.721 & 1.133 & 3.280 & 1.265 & 5.280 & 1.578 & 1.826 & 0.893 & 2.715 & 1.125 & 5.060 & 1.709 & 5.855 & 1.633 \\
			
			& 36 &\best{1.650} &\best{0.821} & 1.725 & 0.872 & \second{1.701} & 0.861 & 1.853 & \second{0.854} & 2.313 & 0.904 & 2.188 & 0.945 & 3.743 & 1.271 & 2.768 & 1.118 & 3.424 & 1.271 & 5.094 & 1.565 & 2.678 & 0.986 & 2.817 & 1.154 & 4.413 & 1.549 & 5.598 & 1.715 \\
			
			& 48 &\best{1.810} &0.921  & 1.937 & 0.937 & 2.153 & 1.041 & \second{1.886} & \best{0.855} & 2.048 & \second{0.886} & 2.231 & 0.956 & 3.853 & 1.306 & 2.637 & 1.088 & 3.009 & 1.520 & 4.884 & 1.530 & 2.584 & 0.937 & 3.038 & 1.199 & 4.109 & 1.473 & 4.795 & 1.568 \\
			
			& 60 &\best{1.850} &\best{0.874}  & 2.128 & 0.999 & 2.064 & 0.953 & \second{1.877} & \second{0.877} & 2.008 & 0.915 & 2.292 & 0.991 & 3.951 & 1.323 & 2.696 & 1.050 & 2.803 & 1.133 & 5.326 & 1.571 & 1.980 & 0.894 & 3.372 & 1.269 & 4.233 & 1.481 & 4.616 & 1.543 \\
			
			\cmidrule(lr){2-30}
			
			& \emph{Avg.}&\best{1.735} &\second{0.854} & 1.861 & 0.924 & \second{1.829} & 0.924 & 1.871 & \best{0.852} & 2.145 & 0.897 & 2.258 &0.957 & 3.749 & 1.284 & 2.705 & 1.097 & 3.129 & 1.297 & 5.123 & 1.561 & 2.267 & 0.927 & 2.985 & 1.186 & 4.453 & 1.553 & 5.216 & 1.614 \\
			\cmidrule(lr){2-30}
			
			\multirow{5}{*}{\rotatebox[origin=c]{90}{ECG}}

			& 96 &\best{0.122} &\best{0.173} &0.185 &\second{0.175} &\second{0.143} &0.179 &0.212 &0.198 &0.159 &0.195 &0.210 &0.212 &\second{0.143} &0.194 &0.214 &0.223 &0.163 &0.183 &0.201 &0.238 &0.261 &0.270 &0.276 &0.264 &0.264 &0.270 &0.255 &0.256 \\
			
			& 192 &\best{0.196} &\best{0.223} &0.235 &\second{0.234} &0.234 &0.245 &0.241 &0.236 &0.237 &0.259 &0.244 &0.255 &\second{0.214} &0.246 &0.237 &0.263 &0.217 &0.237 &0.246 &0.243 &0.283 &0.293 &0.292 &0.306 &0.279 &0.295 &0.273 &0.291\\
			
			& 336 &\best{0.261} &\best{0.277} &0.286 &0.290 &0.288 &0.287 &\second{0.267} &\second{0.270} &0.278 &0.301 &\second{0.267} &0.286 &0.284 &0.286 &\best{0.261} &0.289 &0.284 &0.295 &0.305 &0.310 &0.295 &0.318 &0.300 &0.325 &0.296 &0.317 &0.305 &0.322\\
			
			& 720 &0.314 &\best{0.327} &0.326 &0.341 &0.335 &0.347 &0.329 &0.338 &0.339 &0.356 &\best{0.308} &\second{0.332} &0.338 &0.350 &\second{0.311} &0.342 &0.334 &0.341	&0.359 &0.348 &0.325 &0.342 &0.354 &0.363 &0.328 &0.348 &0.334 &0.359 \\
			
			\cmidrule(lr){2-30}
			
			&\emph{Avg.} &\best{0.225} &\best{0.250} &0.258 &\second{0.260} &0.250 &0.264 &0.262 &\second{0.260} &0.253 &0.277 &0.257 &0.271 &\second{0.244} &0.269 &0.255 &0.279 &0.249 &0.264 &0.277 &0.284 &0.291 &0.305 &0.305 &0.314 &0.291 &0.307 &0.291 &0.307 \\
			\cmidrule(lr){2-30}
			
			\multicolumn{2}{c}{$1^{\text{st}}$ \emph{Count}} & \multicolumn{2}{>{\columncolor{yellow!15}}c}{\best{73}}& \multicolumn{2}{>{\columncolor{yellow!15}}c}{\second{12}} & \multicolumn{2}{>{\columncolor{yellow!15}}c}{0} & \multicolumn{2}{>{\columncolor{yellow!15}}c}{2} & \multicolumn{2}{>{\columncolor{green!15}}c}{0} & \multicolumn{2}{>{\columncolor{green!15}}c}{2} & \multicolumn{2}{>{\columncolor{green!15}}c}{0} & \multicolumn{2}{>{\columncolor{green!15}}c}{1} & \multicolumn{2}{>{\columncolor{green!15}}c}{0} & \multicolumn{2}{>{\columncolor{green!15}}c}{0} & \multicolumn{2}{>{\columncolor{purple!10}}c}{0} & \multicolumn{2}{>{\columncolor{purple!10}}c}{2} & \multicolumn{2}{>{\columncolor{blue!15}}c}{0} & \multicolumn{2}{>{\columncolor{blue!15}}c}{0} \\
			
			\toprule
			
		\end{tabular}
	\end{threeparttable}
		\label{tab::long-term_appendix}
\end{table*}
\begin{table*}[!h]
	\setlength{\tabcolsep}{2pt}
	\scriptsize
	\centering
		\caption{The full results for few-shot forecasting using only 10\% of the training data from the ETT datasets, where the prediction lengths $H \in \{96, 192, 336, 720\}$. The term ``\emph{Avg.}" reports the average result obtained from all four prediction lengths. The best and second best results are highlighted in \best{bold} and \second{underlined}, respectively. The term ``$1^{\text{st}}$ \emph{Count}" indicates the number of times each method achieves the best results.}
	\begin{threeparttable}

		\begin{tabular}{c|c|cc|cc|cc|cc|cc|cc|cc|cc|cc|cc|cc}
			
			\toprule
			\multicolumn{2}{c}{\multirow{1}{*}{\scalebox{1.1}{Models}}} & \multicolumn{2}{c}{LLM-PS}& \multicolumn{2}{c}{CALF} & \multicolumn{2}{c}{TimeLLM} & \multicolumn{2}{c}{GPT4TS} & \multicolumn{2}{c}{PatchTST} & \multicolumn{2}{c}{Crossformer} & \multicolumn{2}{c}{FEDformer} & \multicolumn{2}{c}{TimesNet} & \multicolumn{2}{c}{MICN} & \multicolumn{2}{c}{DLinear} & \multicolumn{2}{c}{TiDE} \\ 
			\multicolumn{2}{c}{} & \multicolumn{2}{c}{\scalebox{0.8}{\textbf{Ours}}}& \multicolumn{2}{c}{\scalebox{0.8}{\citeyearpar{liu2024taming}}} & \multicolumn{2}{c}{\scalebox{0.8}{\citeyearpar{jintime}}} & \multicolumn{2}{c}{\scalebox{0.8}{\citeyearpar{zhou2023onefitsall}}} & \multicolumn{2}{c}{\scalebox{0.8}{\citeyearpar{nie2022pathtst}}} & \multicolumn{2}{c}{\scalebox{0.8}{\citeyearpar{zhang2022crossformer}}} & \multicolumn{2}{c}{\scalebox{0.8}{\citeyearpar{zhou2022fedformer}}} & \multicolumn{2}{c}{\scalebox{0.8}{\citeyearpar{wu2023timesnet}}} & \multicolumn{2}{c}{\scalebox{0.8}{\citeyearpar{wang2022micn}}} & \multicolumn{2}{c}{\scalebox{0.8}{\citeyearpar{zeng2023dlinear}}} & \multicolumn{2}{c}{\scalebox{0.8}{\citeyearpar{das2023tide}}} \\
			
			\cmidrule(lr){3-4} \cmidrule(lr){5-6} \cmidrule(lr){7-8} \cmidrule(lr){9-10} \cmidrule(lr){11-12} \cmidrule(lr){13-14} \cmidrule(lr){15-16} \cmidrule(lr){17-18} \cmidrule(lr){19-20} \cmidrule(lr){21-22} \cmidrule(lr){23-24}
			
			\multicolumn{2}{c}{Metric} & \scalebox{0.8}{MSE} & \scalebox{0.8}{MAE}& \scalebox{0.8}{MSE} & \scalebox{0.8}{MAE} & \scalebox{0.8}{MSE} & \scalebox{0.8}{MAE} & \scalebox{0.8}{MSE} & \scalebox{0.8}{MAE} & \scalebox{0.8}{MSE} & \scalebox{0.8}{MAE} & \scalebox{0.8}{MSE} & \scalebox{0.8}{MAE} & \scalebox{0.8}{MSE} & \scalebox{0.8}{MAE} & \scalebox{0.8}{MSE} & \scalebox{0.8}{MAE} & \scalebox{0.8}{MSE} & \scalebox{0.8}{MAE} & \scalebox{0.8}{MSE} & \scalebox{0.8}{MAE} & \scalebox{0.8}{MSE} & \scalebox{0.8}{MAE} \\
			
			\toprule
			
			\multirow{5}{*}{\rotatebox[origin=c]{90}{ETTm1}} 
			
			& 96  & \best{0.409} & \best{0.411} & \second{0.468} & \second{0.445} & 0.587 & 0.491 & 0.615 & 0.497 & 0.558 & 0.478 & 1.037 & 0.705 & 0.604 & 0.530 & 0.583 & 0.503 & 0.677 & 0.585 & 0.552 & 0.488 & 0.501 & 0.458 \\
			
			& 192 & \best{0.468} & \best{0.440} & \second{0.479} & \second{0.446} & 0.606 & 0.490 & 0.597 & 0.492 & 0.539 & 0.471 & 1.170 & 0.778 & 0.641 & 0.546 & 0.608 & 0.515 & 0.784 & 0.627 & 0.546 & 0.487 & 0.493 & 0.456 \\
			
			& 336 & 0.527 & \second{0.475} & \best{0.499} & \best{0.463} & 0.719 & 0.555 & 0.597 & 0.501 & 0.558 & 0.488 & 1.463 & 0.913 & 0.768 & 0.606 & 0.733 & 0.572 & 0.972 & 0.684 & 0.567 & 0.501 & \second{0.516} & 0.477 \\
			
			& 720 & 0.584 & \second{0.491} & \second{0.572} & 0.496 & 0.632 & 0.514 & 0.623 & 0.513 & 0.574 & 0.498 & 1.693 & 0.997 & 0.771 & 0.606 & 0.768 & 0.548 & 1.449 & 0.800 & 0.606 & 0.522 & \best{0.553} & \best{0.488} \\
			
			\cmidrule(lr){2-24}
			
			& \emph{Avg.}& \best{0.497} & \best{0.454} & \second{0.504} & \second{0.462} & 0.636 & 0.512 & 0.608 & 0.500 & 0.557 & 0.483 & 1.340 & 0.848 & 0.696 & 0.572 & 0.673 & 0.534 & 0.970 & 0.674 & 0.567 & 0.499 & 0.515 & 0.469 \\
			
			\midrule
			
			\multirow{5}{*}{\rotatebox[origin=c]{90}{ETTm2}} 
			
			& 96 & \best{0.186} & \best{0.263}  & 0.190 & 0.268 & 0.189 & 0.270 & \second{0.187} & \second{0.266} & 0.189 & 0.268 & 1.397 & 0.866 & 0.222 & 0.314 & 0.214 & 0.288 & 0.389 & 0.448 & 0.225 & 0.320 & 0.191 & 0.269 \\
			
			& 192 & \best{0.239} & \best{0.297} & 0.257 & 0.311 & 0.264 & 0.319 & 0.253 & 0.308 & \second{0.248} & \second{0.307} & 1.757 & 0.987 & 0.284 & 0.351 & 0.271 & 0.325 & 0.622 & 0.575 & 0.291 & 0.362 & 0.256 & 0.310 \\
			
			& 336& \best{0.308} & \second{0.344}  & 0.323 & \best{0.334} & 0.327 & 0.358 & 0.332 & 0.353 & \second{0.311} & 0.346 & 2.075 & 1.086 & 0.392 & 0.419 & 0.329 & 0.356 & 1.055 & 0.755 & 0.354 & 0.402 & 0.321 & 0.349 \\
			
			& 720 & \best{0.389} & \best{0.390} & 0.441 & \second{0.410} & 0.454 & 0.428 & 0.438 & 0.417 & \second{0.435} & 0.418 & 2.712 & 1.253 & 0.527 & 0.485 & 0.473 & 0.448 & 2.226 & 1.087 & 0.446 & 0.447 & 0.446 & 0.421 \\
			
			\cmidrule(lr){2-24}
			
			& \emph{Avg.} & \best{0.281} & \best{0.324} & 0.302 & \second{0.330} & 0.308 & 0.343 & 0.303 & 0.336 & \second{0.295} & 0.334 & 1.985 & 1.048 & 0.356 & 0.392 & 0.321 & 0.354 & 1.073 & 0.716 & 0.329 & 0.382 & 0.303 & 0.337 \\
			
			\midrule
			
			\multirow{5}{*}{\rotatebox[origin=c]{90}{ETTh1}} 
			
			& 96 & 0.586 & 0.529  & 0.468 & 0.457 & 0.500 & 0.464 & \second{0.462} & \second{0.449} & \best{0.433} & \best{0.428} & 1.129 & 0.775 & 0.651 & 0.563 & 0.855 & 0.625 & 0.689 & 0.592 & 0.590 & 0.515 & 0.642 & 0.545 \\
			
			& 192 & 0.620 & 0.537 & \second{0.550} & 0.501 & 0.590 & 0.516 & 0.551 & \second{0.495} & \best{0.509} & \best{0.474} & 1.832 & 0.922 & 0.666 & 0.562 & 0.791 & 0.589 & 1.160 & 0.748 & 0.634 & 0.541 & 0.761 & 0.595 \\
			
			& 336 & 0.658 & 0.553 & \second{0.581} & \second{0.521} & 0.638 & 0.542 & 0.630 & 0.539 & \best{0.572} & \best{0.509} & 2.022 & 0.973 & 0.767 & 0.602 & 0.939 & 0.648 & 1.747 & 0.899 & 0.659 & 0.554 & 0.789 & 0.610 \\
			
			& 720 & \best{0.664} & \best{0.563} & 0.978 & 0.685 & 1.334 & 0.816 & 1.113 & 0.738 & 1.221 & 0.773 & 1.903 & 0.986 & 0.918 & 0.703 & 0.876 & 0.641 & 2.024 & 1.019 & \second{0.708} & \second{0.598} & 0.927 & 0.667 \\
			
			\cmidrule(lr){2-24}
			
			& \emph{Avg.} & \best{0.632} & \second{0.546}& \second{0.644} & \best{0.541} & 0.765 & 0.584 & 0.689 & 0.555 & 0.683 & 0.645 & 1.744 & 0.914 & 0.750 & 0.607 & 0.865 & 0.625 & 1.405 & 0.814 & 0.647 & 0.552 & 0.779 & 0.604 \\
			
			\midrule
			
			\multirow{5}{*}{\rotatebox[origin=c]{90}{ETTh2}} 
			
			& 96 & 0.332 & 0.372  & \best{0.314} & 0.360 & 0.329 & 0.365 & \second{0.327} & \second{0.359} & \best{0.314} & \best{0.354} & 2.482 & 1.206 & 0.359 & 0.404 & 0.372 & 0.405 & 0.510 & 0.502 & 0.361 & 0.407 & 0.337 & 0.379 \\
			
			& 192 & \best{0.398} & 0.412 & 0.404 & \second{0.411} & 0.414 & 0.413 & \second{0.403} & \best{0.405} & 0.420 & 0.415 & 3.136 & 1.372 & 0.460 & 0.461 & 0.483 & 0.463 & 1.809 & 1.036 & 0.444 & 0.453 & 0.424 & 0.427 \\
			
			& 336 & \best{0.430} & \second{0.431} & 0.458 & 0.452 & 0.579 & 0.506 & 0.568 & 0.499 & 0.543 & 0.489 & 2.925 & 1.331 & 0.569 & 0.530 & 0.541 & 0.496 & 3.250 & 1.419 & 0.509 & 0.501 & \second{0.435} & \best{0.426} \\
			
			& 720 & \second{0.476} & \best{0.463} & 0.502 & 0.487 & 1.034 & 0.711 & 1.020 & 0.725 & 0.926 & 0.691 & 4.014 & 1.603 & 0.827 & 0.707 & 0.510 & 0.491 & 4.564 & 1.676 & \best{0.453} & \second{0.471} & 0.489 & 0.480 \\
			
			\cmidrule(lr){2-24}
			
			& \emph{Avg.}& \best{0.409} & \best{0.420} & \second{0.419} & \second{0.427} & 0.589 & 0.498 & 0.579 & 0.497 & 0.550 & 0.487 & 3.139 & 1.378 & 0.553 & 0.525 & 0.476 & 0.463 & 2.533 & 1.158 & 0.441 & 0.458 & 0.421 & 0.428 \\
			
			\midrule
			
			\multicolumn{2}{c}{$1^{\text{st}}$ \emph{Count}}& \multicolumn{2}{c}{\best{23}} & \multicolumn{2}{c}{5} & \multicolumn{2}{c}{0} & \multicolumn{2}{c}{1} & \multicolumn{2}{c}{\second{8}} & \multicolumn{2}{c}{0} & \multicolumn{2}{c}{0} & \multicolumn{2}{c}{0} & \multicolumn{2}{c}{0} & \multicolumn{2}{c}{1} & \multicolumn{2}{c}{3} \\
			
			\toprule
			
		\end{tabular}
	\end{threeparttable}
\label{tab::few-short}
\end{table*}
\begin{table*}[!htb]
	\setlength{\tabcolsep}{2pt}
	\scriptsize
	\centering
\caption{Full results for zero-shot forecasting on the ETT datasets, where prediction lengths $H \in \{96, 192, 336, 720\}$. ``h1", ``h2", ``m1", and ``m2" denote ETTh1, ETTh2, ETTm1, and ETTm2 respectively. ``h1 $\to$ m1" indicates that models trained on ETTh1 are evaluated on ETTm1,  and the same applies to other items. The team ``\emph{Avg.}" reports the results averaged from all four prediction lengths. The best and the second best results are in \best{bold} and \second{underlined}. ``$1^{\text{st}}$ \emph{Count}'' indicates the number of times each method achieves the best results.}
	\begin{threeparttable}
		\begin{tabular}{c|c|cc|cc|cc|cc|cc|cc|cc|cc|cc|cc|cc}
			
			\toprule
			\multicolumn{2}{c}{\multirow{2}{*}{\scalebox{1.1}{Models}}} & \multicolumn{2}{c}{LLM-PS}& \multicolumn{2}{c}{CALF} & \multicolumn{2}{c}{TimeLLM} & \multicolumn{2}{c}{GPT4TS} & \multicolumn{2}{c}{PatchTST} & \multicolumn{2}{c}{Crossformer} & \multicolumn{2}{c}{FEDformer} & \multicolumn{2}{c}{TimesNet} & \multicolumn{2}{c}{MICN} & \multicolumn{2}{c}{DLinear} & \multicolumn{2}{c}{TiDE} \\ 
			\multicolumn{2}{c}{} & \multicolumn{2}{c}{\scalebox{0.8}{\textbf{Ours}}}& \multicolumn{2}{c}{\scalebox{0.8}{\citeyearpar{liu2024taming}}} & \multicolumn{2}{c}{\scalebox{0.8}{\citeyearpar{jintime}}} & \multicolumn{2}{c}{\scalebox{0.8}{\citeyearpar{zhou2023onefitsall}}} & \multicolumn{2}{c}{\scalebox{0.8}{\citeyearpar{nie2022pathtst}}} & \multicolumn{2}{c}{\scalebox{0.8}{\citeyearpar{zhang2022crossformer}}} & \multicolumn{2}{c}{\scalebox{0.8}{\citeyearpar{zhou2022fedformer}}} & \multicolumn{2}{c}{\scalebox{0.8}{\citeyearpar{wu2023timesnet}}} & \multicolumn{2}{c}{\scalebox{0.8}{\citeyearpar{wang2022micn}}} & \multicolumn{2}{c}{\scalebox{0.8}{\citeyearpar{zeng2023dlinear}}} & \multicolumn{2}{c}{\scalebox{0.8}{\citeyearpar{das2023tide}}} \\
			
			\cmidrule(lr){3-4} \cmidrule(lr){5-6} \cmidrule(lr){7-8} \cmidrule(lr){9-10} \cmidrule(lr){11-12} \cmidrule(lr){13-14} \cmidrule(lr){15-16} \cmidrule(lr){17-18} \cmidrule(lr){19-20} \cmidrule(lr){21-22} \cmidrule(lr){23-24}
			
			\multicolumn{2}{c}{Metric} & \scalebox{0.8}{MSE} & \scalebox{0.8}{MAE} & \scalebox{0.8}{MSE} & \scalebox{0.8}{MAE} & \scalebox{0.8}{MSE} & \scalebox{0.8}{MAE} & \scalebox{0.8}{MSE} & \scalebox{0.8}{MAE} & \scalebox{0.8}{MSE} & \scalebox{0.8}{MAE} & \scalebox{0.8}{MSE} & \scalebox{0.8}{MAE} & \scalebox{0.8}{MSE} & \scalebox{0.8}{MAE} & \scalebox{0.8}{MSE} & \scalebox{0.8}{MAE} & \scalebox{0.8}{MSE} & \scalebox{0.8}{MAE} & \scalebox{0.8}{MSE} & \scalebox{0.8}{MAE} & \scalebox{0.8}{MSE} & \scalebox{0.8}{MAE} \\
			
			\toprule
			
			\multirow{5}{*}{\rotatebox[origin=c]{90}{h1 $\rightarrow$ m1}} 
			
			& 96   & \best{0.719} & 0.575 & 0.767 & 0.564 & 0.804 & 0.565 & 0.809 & 0.563 & 0.908 & 0.596 & 0.856 & 0.649 & \second{0.731} & 0.561 & 0.764 & 0.563 & 0.832 & 0.621 & 0.735 & \second{0.554} & 0.748 & \best{0.551} \\
			
			& 192 & \best{0.724} & \best{0.528} & 0.753 & 0.570 & 0.827 & 0.593 & 0.799 & 0.567 & 0.927 & 0.616 & 0.906 & 0.684 & \second{0.746} & 0.573 & 0.798 & \second{0.562} & 1.288 & 0.854 & 0.752 & 0.570 & 0.779 & 0.571 \\
			
			& 336 & \best{0.725} & \best{0.543}  & \second{0.745} & \second{0.575} & 0.835 & 0.600 & 0.803 & 0.577 & 0.920 & 0.621 & 1.104 & 0.796 & 0.775 & 0.596 & 0.790 & 0.584 & 1.721 & 0.972 & 0.749 & 0.579 & 0.775 & 0.580 \\
			
			& 720 & \best{0.730} & \best{0.564} & \second{0.758} & 0.590 & 0.922 & 0.644 & 0.783 & \second{0.589} & 0.822 & 0.608 & 1.131 & 0.816 & 0.808 & 0.625 & 0.827 & 0.594 & 1.915 & 1.036 & 0.805 & 0.606 & 0.795 & 0.595 \\
			
			\cmidrule(lr){2-24}
			
			& \emph{Avg.} & \best{0.721} & \best{0.541} & \second{0.755} & \second{0.574} & 0.847 & 0.600 & 0.798 & \second{0.574} & 0.894 & 0.610 & 0.999 & 0.736 & 0.765 & 0.588 & 0.794 & 0.575 & 1.439 & 0.870 & 0.760 & 0.577 & 0.774 & \second{0.574} \\
			
			\midrule
			
			\multirow{5}{*}{\rotatebox[origin=c]{90}{h1 $\rightarrow$ m2}} 
			
			& 96  & 0.217 & 0.327 & 0.218 & 0.301 & \best{0.212} & \best{0.298} & 0.218 & 0.304 & 0.219 & 0.305 & 0.611 & 0.588 & 0.257 & 0.345 & 0.245 & 0.322 & 0.496 & 0.556 & 0.239 & 0.343 & \second{0.215} & \second{0.299} \\
			
			& 192 & 0.289 & 0.340 & \second{0.278} & \best{0.334} & \best{0.277} & 0.338 & 0.279 & 0.338 & 0.280 & 0.341 & 0.789 & 0.685 & 0.318 & 0.380 & 0.293 & 0.346 & 1.798 & 1.137 & 0.320 & 0.397 & \best{0.277} & \second{0.335} \\
			
			& 336 & \best{0.330} & \best{0.364} & 0.338 & \second{0.369} & \second{0.336} & 0.371 & 0.342 & 0.376 & 0.341 & 0.376 & 1.469 & 0.927 & 0.375 & 0.417 & 0.361 & 0.382 & 2.929 & 1.472 & 0.409 & 0.453 & 0.337 & 0.370 \\
			
			& 720 & \best{0.429} & \best{0.411} & \second{0.431} & \second{0.418} & 0.435 & 0.424 & \second{0.431} & 0.419 & 0.432 & 0.426 & 1.612 & 0.957 & 0.480 & 0.472 & 0.460 & 0.432 & 4.489 & 1.782 & 0.629 & 0.565 & \best{0.429} & \second{0.418} \\
			
			\cmidrule(lr){2-24}
			
			& \emph{Avg.}& 0.316 & 0.361 & 0.316 & \best{0.355} & \second{0.315} & \second{0.357} & 0.317 & 0.359 & 0.318 & 0.362 & 1.120 & 0.789 & 0.357 & 0.403 & 0.339 & 0.370 & 2.428 & 1.236 & 0.399 & 0.439 & \best{0.314} & \best{0.355} \\
			
			\midrule
			
			\multirow{5}{*}{\rotatebox[origin=c]{90}{h2 $\rightarrow$ m1}} 
			
			& 96 & \best{0.684} & \best{0.538}  & 0.897 & 0.589 & 0.891 & 0.587 & 0.985 & 0.604 & 0.815 & \second{0.560} & 1.032 & 0.620 & \second{0.734} & 0.578 & 1.205 & 0.678 & 0.743 & 0.577 & 0.762 & 0.567 & 0.819 & 0.566 \\
			
			& 192 & \best{0.702} & \best{0.541} & 0.864 & 0.584 & 0.850 & \second{0.583} & 0.872 & 0.600 & 0.900 & 0.606 & 1.176 & 0.676 & \second{0.723} & 0.594 & 1.159 & 0.670 & 0.750 & 0.588 & 0.785 & 0.588 & 0.845 & 0.586 \\
			
			& 336 & \best{0.738} & \best{0.569} & 0.816 & \second{0.585} & 0.853 & 0.594 & 0.926 & 0.614 & 0.906 & 0.602 & 1.199 & 0.718 & \second{0.750} & 0.590 & 1.197 & 0.689 & 0.764 & 0.606 & 0.767 & 0.594 & 0.834 & 0.595 \\
			
			& 720 & \best{0.735} & \best{0.558} & 0.768 & \second{0.589} & 0.879 & 0.616 & 0.899 & 0.624 & 0.866 & 0.619 & 1.373 & 0.832 & \second{0.760} & 0.592 & 1.583 & 0.784 & 0.801 & 0.634 & 0.800 & 0.627 & 0.867 & 0.616 \\
			
			\cmidrule(lr){2-24}
			
			& \emph{Avg.} & \best{0.714} & \best{0.552} & 0.836 & \second{0.586} & 0.868 & 0.595 & 0.920 & 0.610 & 0.871 & 0.596 & 1.195 & 0.711 & \second{0.741} & 0.588 & 1.286 & 0.705 & 0.764 & 0.601 & 0.778 & 0.594 & 0.841 & 0.590 \\
			
			\midrule
			
			\multirow{5}{*}{\rotatebox[origin=c]{90}{h2 $\rightarrow$ m2}}
			
			& 96  & 0.231 & 0.315 & \best{0.225} & \best{0.310} & 0.228 & \second{0.311} & 0.235 & 0.316 & 0.288 & 0.345 & 0.821 & 0.634 & 0.261 & 0.347 & 0.244 & 0.324 & 0.327 & 0.414 & 0.264 & 0.366 & \second{0.226} & 0.315 \\
			
			& 192 & \second{0.284} & \best{0.338} & \best{0.283} & 0.342 & \best{0.283} & \second{0.341} & 0.287 & 0.346 & 0.344 & 0.375 & 1.732 & 1.018 & 0.313 & 0.370 & 0.331 & 0.374 & 0.450 & 0.485 & 0.394 & 0.452 & 0.289 & 0.348 \\
			
			& 336 & \best{0.338} & \best{0.369} & 0.340 & 0.373 & 0.343 & 0.376 & 0.361 & 0.391 & 0.438 & 0.425 & 2.587 & 1.393 & 0.401 & 0.431 & 0.386 & 0.405 & 0.526 & 0.526 & 0.506 & 0.513 & \second{0.339} & \second{0.372} \\
			
			& 720 & \second{0.433}& \second{0.419} & \best{0.429} & \best{0.418} & 0.437 & 0.424 & 0.444 & 0.433 & 0.611 & 0.588 & 3.034 & 1.452 & 0.487 & 0.472 & 0.485 & 0.458 & 0.806 & 0.652 & 0.822 & 0.655 & \second{0.433} & 0.422 \\
			
			\cmidrule(lr){2-24}
			
			& \emph{Avg.} & 0.322 & \best{0.359}& \best{0.319} & \second{0.360} & 0.322 & 0.363 & 0.331 & 0.371 & 0.420 & 0.433 & 2.043 & 1.124 & 0.365 & 0.405 & 0.361 & 0.390 & 0.527 & 0.519 & 0.496 & 0.496 & \second{0.321} & 0.364 \\
			
			\midrule
			
			\multicolumn{2}{c}{$1^{\text{st}}$ \emph{Count}}& \multicolumn{2}{c}{\best{27}} & \multicolumn{2}{c}{\second{8}} & \multicolumn{2}{c}{4} & \multicolumn{2}{c}{0} & \multicolumn{2}{c}{0} & \multicolumn{2}{c}{0} & \multicolumn{2}{c}{0} & \multicolumn{2}{c}{0} & \multicolumn{2}{c}{0} & \multicolumn{2}{c}{0} & \multicolumn{2}{c}{5} \\
			
			\toprule
			
		\end{tabular}
	\end{threeparttable}
			\label{tab::zero-short}
\end{table*}
\subsection{Baseline Time Series Forecasting Methods}
\label{sec_apx_btsfm}
In this paper, we compare an extensive range of SOTA methods, primarily categorized as follows: \\ \\
1) \textbf{LLM-based Methods}: 
\begin{itemize}[itemsep=-0.1em]
	\item GPT4TS~\cite{zhou2023onefitsall}, the pioneering work that employs LLMs for time series forecasting by segmenting continuous time series into discrete tokens compatible with LLMs.
	\item TimeLLM~\cite{jintime}, which proposes patch reprogramming to encode prior knowledge from time series datasets into prompts for guiding the LLM in time series forecasting.
	\item CALF~\cite{liu2024taming} trains separate branches for temporal and textual modalities and closely aligns them with leveraging textual knowledge in LLMs for time series prediction.
\end{itemize}
2) \textbf{Transformer-based Methods}: 
\begin{itemize}[itemsep=-0.1em]
	\item Crossformer~\cite{zhang2022crossformer} identifies that relationships between variables in time series data are crucial for time series forecasting, so it captures them using attention mechanisms.
	\item FEDformer~\cite{zhou2022fedformer} and Autoformer~\cite{wu2021autoformer}, which decouple seasonal and trend components in the frequency domain and learn them based on attention mechanisms.
	\item PatchTST~\cite{nie2022pathtst}, the first work proposed partitioning input series into multiple patches, effectively enhancing the long-range time series prediction capability of Transformers.
	\item iTransformer~\cite{liu2023itransformer} captures relationships between variables by transposing input time series data.
	\item ETSformer~\cite{woo2022etsformer} introduces both smoothing attention and frequency attention to replace the original self-attention mechanism in Transformers.
\end{itemize}
3) \textbf{CNN-based Methods}: 
\begin{itemize}[itemsep=-0.1em]
	\item TimesNet~\cite{wu2023timesnet}, which selects representative periods in the frequency domain and processes them using 2D convolution layers.
	\item TCN~\cite{BaiTCN2018} conducts a systematic evaluation of generic convolutional and recurrent architectures for sequence modeling.
	\item MICN~\cite{wang2022micn} decomposes the time series signal into seasonal and trend components and learns them separately using convolutional and linear regression layers.
\end{itemize}
4) \textbf{MLP-based Methods}:
\begin{itemize}[itemsep=-0.1em]
	\item DLinear~\cite{zeng2023dlinear} explores the application of linear layers in time series tasks and achieves efficient time series prediction.
	\item TiDE~\cite{das2023tide} designs an encoder-decoder structure based on MLP, which can achieve comparable performance with Transformers while requiring less computation burdens.
	\item TimeMixer~\cite{wangtimemixer} downsamples the time series signal into multiple-scale inputs for ensemble predictions in the MLP model.	
\end{itemize}
Additionally, we compare with early time series forecasting methods:
\begin{itemize}[itemsep=-0.1em]
	\item N-BEATS \cite{oreshkin2019nbeats} is the first to apply deep learning models to time series forecasting tasks and designed a deep model based on residual links and Fourier series, achieving better performance than traditional statistical methods.
	\item N-HiTS \cite{challu2022nhits} finds that N-BEATS becomes slow as the forecast length increases and reduces the time series data length by downsampling the input series, thereby effectively improving the model's inference speed.
\end{itemize}
\subsection{Benchmark Datasets}
\label{sec_apx_bd}
\begin{figure*}[!t]
	\centering
	\includegraphics[scale=0.335]{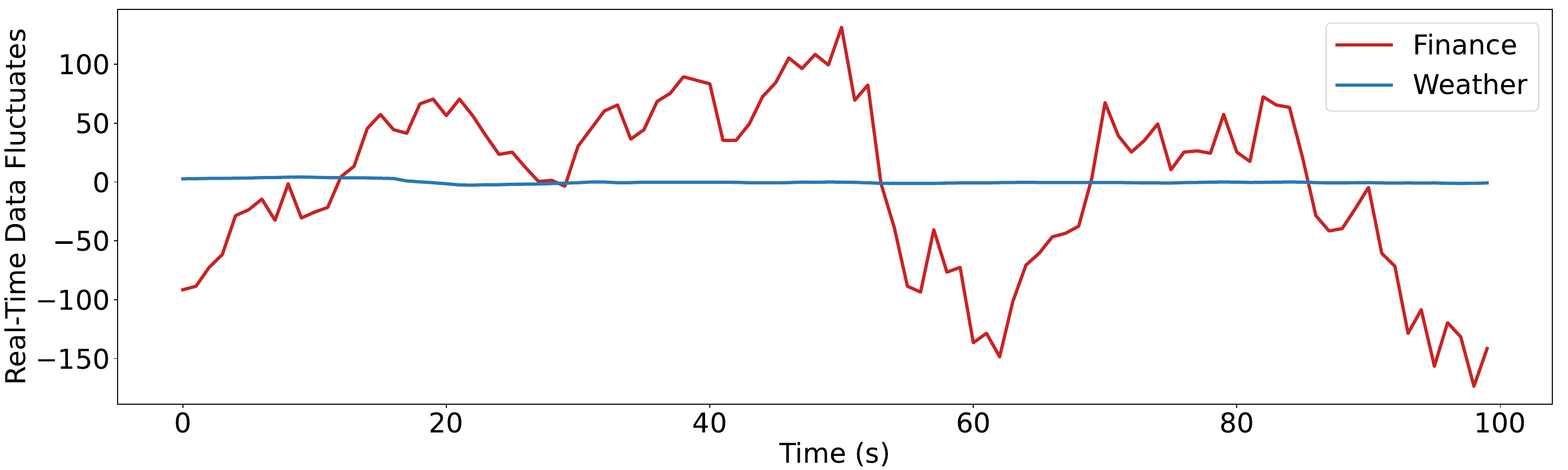}
	\caption{Visualization of the time series belongs to the Weather dataset and the finance subset in the M4 dataset. The temperature readings measured by the meteorological station are generally stable, whereas stock prices in financial markets fluctuate rapidly around the average value.}
	\label{fig_vistime}
\end{figure*}

In this paper, we evaluate our proposed LLM-PS on the Electricity Transformer Temperature (ETT)~\cite{haoyietalinformer}, Weather~\cite{wu2021autoformer}, Traffic~\cite{wu2021autoformer}, Illness (ILI),~\cite{wu2021autoformer}, Electricity~\cite{electricity}, M4~\cite{makridakis2018m4}, and Electrocardiography (ECG)~\cite{moody2001impact} datasets. These datasets are sourced from different domains, such as finance and meteorology. In general, the time series in these datasets exhibit distinct patterns, as shown in Fig.~\ref{fig_vistime}. The details of these time series datasets are as follows.\\
\textbf{ETT} records measurements from an electricity transformer over an extended period, primarily focusing on seven variables, including the target variable ``oil temperature" and six power load features. The ETT dataset contains four subsets (ETTh1, ETTh2, ETTm1, and ETTm2) sampled at different frequencies (hourly and minutely) from various locations. The training, validation, and test sets of ETT contain data sampled over 12, 4, and 4 months, respectively.\\
\textbf{Weather} is a meteorological dataset used for climate modeling and environmental research, which records 21 meteorological indicators every 10 minutes throughout the entire year of 2020, such as air temperature and humidity.\\
\textbf{Traffic} is commonly employed for traffic flow forecasting, which predicts the spatio-temporal traffic volume by considering historical traffic data and additional features from adjacent locations. This dataset captures traffic volume measurements every 15 minutes at 862 sensor locations situated along two main highways from July 1, 2016, to July 2, 2018.\\
\textbf{Illness} includes weekly data from the Centers for Disease Control and Prevention in the United States from 2002 to 2021, recording the ratio of patients with influenza to the total number of patients. Illness contains seven variables, such as age and number of providers.\\
\textbf{Electricity} consists of 321 variables related to energy utilities for energy production and usage in the United States from over 2,000 U.S. utilities in 2017.\\
\textbf{M4} comprises 100,000 time series utilized in the fourth edition of the Makridakis Forecasting Competition. This dataset encompasses yearly, quarterly, monthly, and various other datasets.\\
\textbf{ECG} is a collection of biomedical datasets primarily used for research in electrocardiogram (ECG) analysis. This dataset contains 48 half-hour two-channel ambulatory ECG recordings from 47 subjects with a sampling rate of 360 Hz.
\subsection{Metrics of Time Series Forecasting}
\label{sec_apx_metrics}
In this paper, we mainly employ five widely used metrics to assess model performance, including Mean Squared Error (MSE), Mean Absolute Error (MAE), Mean Absolute Scaled Error (MSAE), Symmetric Mean Absolute Percentage Error (SMAPE), and Overall Weighted Average (OWA).\\
\textbf{MSE} measures the average of the squared differences between the predicted and actual values. MSE gives more weight to more significant errors because the errors are squared, thereby sensitive to outliers. Given $T$ time steps ground-truth time series signal $\mathbf{Y}=\left\{\mathbf{x}_{H+1}, \ldots, \mathbf{x}_{H+T}\right\}\in \mathbb{R}^{T \times V}$ and prediction $\hat{\mathbf{Y}}=\left\{\hat{\mathbf{x}}_{H+1}, \ldots, \hat{\mathbf{x}}_{H+T}\right\}\in \mathbb{R}^{T \times V}$, MSE is calculated as:
\begin{equation}
	\operatorname{MSE}=\frac{1}{T} \sum_{i=1}^{T}\left(\mathbf{x}_{H+i}-\hat{\mathbf{x}}_{H+i}\right)^2.
\end{equation}
\textbf{MAE} quantifies the average squared differences between predicted and actual values. It is less sensitive to outliers than MSE because it does not square the errors, treating all errors linearly. MAE is computed as:
\begin{equation}
	\operatorname{MAE}=\frac{1}{T}\sum_{i=1}^{T}\left|\mathbf{x}_{H+i}-\hat{\mathbf{x}}_{H+i}\right|.
\end{equation}
\textbf{MSAE} evaluates the accuracy of a model by scaling the absolute error relative to the actual values, which is calculated as:
\begin{equation}
	\text{MSAE} = \frac{1}{T} \sum_{i=1}^{T} \frac{|\mathbf{x}_{H+i} - \hat{\mathbf{x}}_{H+i}|}{|\mathbf{x}_{H+i}|}.
\end{equation}
\textbf{SMAPE} aims to provide a more balanced error with a symmetry formula, especially when the actual values approach zero. This helps mitigate the instability in MAPE when the actual values are small. SMAPE is calculated as:
\begin{equation}
	\text{SMAPE} = \frac{1}{T} \sum_{i=1}^{T} \frac{2 |\mathbf{x}_{H+i} - \hat{\mathbf{x}}_{H+i}|}{\mathbf{x}_{H+i} + |\hat{\mathbf{x}}_{H+i}|}.
\end{equation}
\textbf{OWA} is commonly used for multiple tasks or criteria, and the importance of each task or metric differs. It allows for a balanced evaluation by adjusting the contribution of each task based on its relative importance. OWA is computed as:
\begin{equation}
	\text{OWA} = \sum_{i=1}^{N} w_i \cdot v_i,
\end{equation}
where $N$ is the number of tasks (metrics), $w_i$ is the weight assigned to the $i$-th task (metric) and $\sum_{i=1}^N w_i=1$, $v_i$ is the evaluation result of the $i$-th task (metric).
\section{Model Analysis}
\label{sec_apx_ma}
\begin{figure*}[!t]
	\centering
	\includegraphics[scale=0.565]{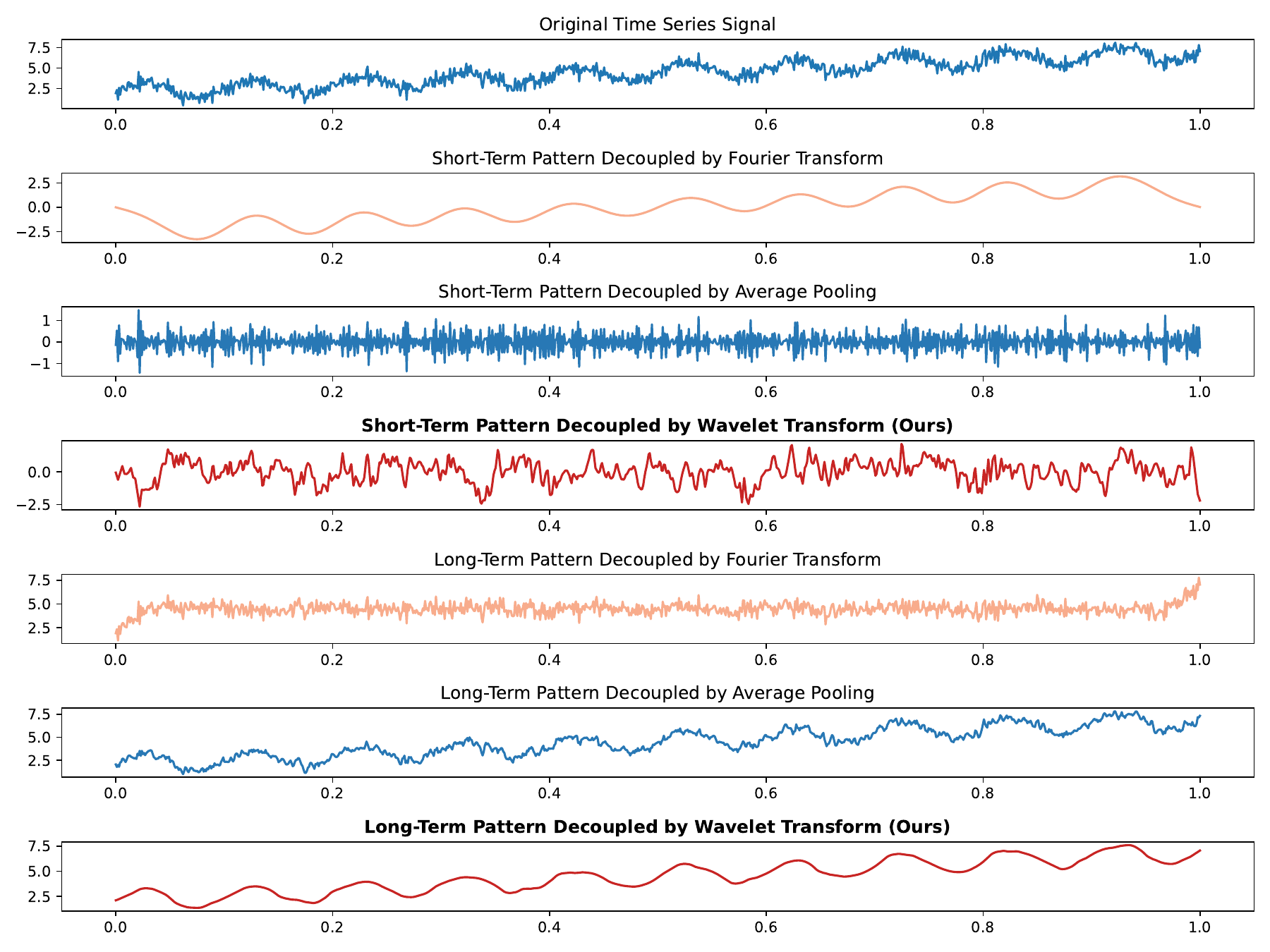}
	\caption{Visualization results of temporal patterns decoupled by our proposed wavelet-transform-based decoupling technique and other widely-used decoupling methods based on Fourier transform and average pooling.}
	\label{fig_visdtp}
\end{figure*}
\subsection{Wavelet Transform}
\label{sec_apx_wt}
Our MSCNN employs the DauBechies 4 (DB4) wavelet transform with four vanishing moments to decouple low-frequency components $\mathbf{W}^{b}_{\text{low}}$ and high-frequency components $\{\mathbf{W}^{b}_{\text{high}\_i}\}_{i=1}^{w}$. The DB4 wavelet transform is widely used in time series signal processing due to its compact support and smoothness. Specifically, the DB4 wavelet transform decomposes the input time series via sequential decomposition and downsampling steps. In the decomposition step, the time series $\mathbf{X}=\left\{\mathbf{x}_1, \ldots, \mathbf{x}_H\right\} \in \mathbb{R}^{H \times V}$ is progressively decomposed into multiple frequency bands with $w$ levels, encompassing both low-frequency and high-frequency components. At $i$-th decomposition level, the time series $\mathbf{X}$ is passed through low-pass filter $f^{i}_\text{low}$ and high-pass filter $f^{i}_\text{high}$ to generate the low-frequency approximation coefficients $\mathbf{a}_i$ and high-frequency detail coefficients $\mathbf{d}_i$: 
\begin{equation}
	\mathbf{a}_i[n] =\sum_{k=0}^{L-1} f^{i}_\text{low}[k] \cdot \mathbf{X}[2 n-k], \quad
	\mathbf{d}_i[n] =\sum_{k=0}^{L-1} f^{i}_\text{high}[k] \cdot \mathbf{X}[2 n-k], \quad n=1,2, \ldots,\left\lfloor\frac{H+L-1}{2}\right\rfloor,
\end{equation}
where $L=8$ denotes the filter length. After $w$ decomposition levels, the outputs from these filters undergo downsampling by a factor of 2. The final approximation $\mathbf{a}_{w}$, which represents the smoothest features of the signal, serves as the low-frequency components $\mathbf{W}^{b}_{\text{low}}$. Meanwhile, the detail coefficients $\{\mathbf{d}_{i}\}_{i=1}^{w}$, which capture diverse high-frequency details or changes in the input time series, constitute the high-frequency components $\{\mathbf{W}^{b}_{\text{high}\_i}\}_{i=1}^{w}$.

Conversely, the inverse DB4 wavelet transform reconstructs the original signal through upsampling and reversed-order filtering. During the upsampling step, the approximation coefficients and detail coefficients are expanded by a factor of 2 to recover the original signal length. In the filtering step, the upsampled coefficients are convolved with the same low-pass and high-pass filters used during the decomposition in reverse order to reconstruct the time series signal:
\begin{equation}
	\mathbf{X}=\sum_{i=1}^w\left(\tilde{\mathbf{a}}_i+\tilde{\mathbf{d}}_i\right).
\end{equation}
Here, $\tilde{\mathbf{a}}_i$ and $\tilde{\mathbf{d}}_i$ represent reconstruction coefficients and details, which are calculated as follows:
\begin{equation}
	\tilde{\mathbf{a}}_i[m]=\sum_n \tilde{f}^{i}_\text{low}[m-2 n] \cdot \mathbf{a}_i[n], \quad
	\tilde{\mathbf{d}}_i[m]=\sum_n \tilde{f}^{i}_\text{high}[m-2 n] \cdot \mathbf{d}_i[n], \quad m=1,2, \ldots, H.
\end{equation}
$\tilde{f}^{i}_\text{low}$ (low-pass) and $\tilde{f}^{i}_\text{high}$ (high-pass), derived as time-reversed versions of the decomposition filter $f^{i}_\text{low}$ (low-pass) and $f^{i}_\text{high}$, as follows:
\begin{equation}
	\tilde{f}^{i}_\text{low}[k]=f^{i}_\text{low}[L-1-k], \quad \tilde{f}^{i}_\text{high}[k]=f^{i}_\text{high}[L-1-k], \quad k=0,1, \ldots, L-1.
\end{equation}
\subsection{Visualizations of Decoupled Temporal Patterns}
\label{sec_apx_visdtp}
In this section, we compare our proposed wavelet-transform-based temporal pattern decoupling technique with other methods that employ Fourier transform and average pooling. Fig.~\ref{fig_visdtp} visualizes the decomposed short-term and long-term patterns. We can observe that our decoupling method achieves significantly better results than other decoupling methods. In particular, our method accurately captures short-term patterns that mirror the periodic fluctuations of the original time series, while effectively representing the overall trend in the long-term patterns.
\subsection{Receptive Field of MSCNN}
\label{sec_apx_rf}
In CNNs, the receptive field refers to the region of the input associated with each output in a convolutional layer. The size of the receptive field determines how much of the input the network observes during convolution operations. In general, the receptive field of the output for a convolutional layer depends on the kernel size $k$, stride $s$, and the receptive field of the input $R_{\text {in}}$, which is formulated as: 
\begin{equation}
	R_{\text {out}}=R_{\text {in}}+(k-1) \times s.
\end{equation}

In our MSCNN block, the features are divided into multiple branches and processed by parallel 3$\times$3 convolutional layers with an identical stride $s$=1. Given the features $\{\bar{\mathbf{F}}_{1}, \ldots, \bar{\mathbf{F}}_{B}\}$ obtained from $B$ branches in Eq.~\eqref{eq_mscnn}, we assume that the receptive field of input features $R_{\text {in}}$=1 and $B$=4, and the receptive fields of multiple-branch features are $[3, 5, 7, 9]$. Here, features with a small receptive field focus on short-term patterns, namely the periodic fluctuations, while features with a large receptive field focus on long-term patterns, namely the overall trend. Therefore, our proposed MSCNN can effectively capture both short-term and long-term patterns.
\begin{figure*}[!t]
	\centering
	\includegraphics[scale=0.55]{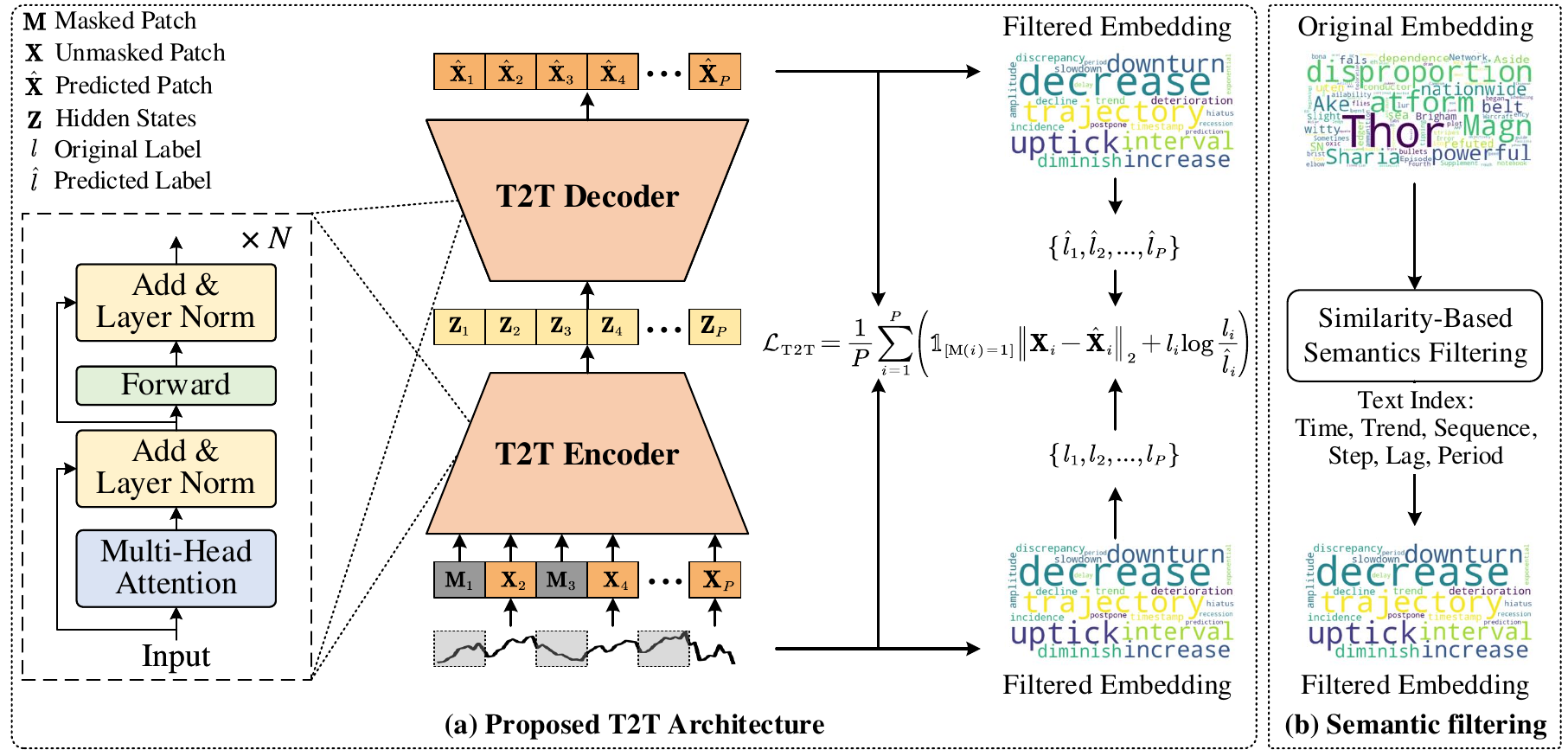}
	\caption{The diagram of our proposed Time-to-Text (T2T) module. The input time series is first divided into $P$ patches, with some patches randomly masked. These patches are then fed into the encoder and decoder to facilitate T2T's accurate reconstruction and encourage T2T to learn meaningful semantic information by encouraging it to accurately predict the semantic labels of the time patches.}
	\label{fig_t2t}
\end{figure*}
\begin{wrapfigure}{R}{0.4\textwidth}
	\vspace{-4mm}
	\centering
	\includegraphics[width=5.8cm]{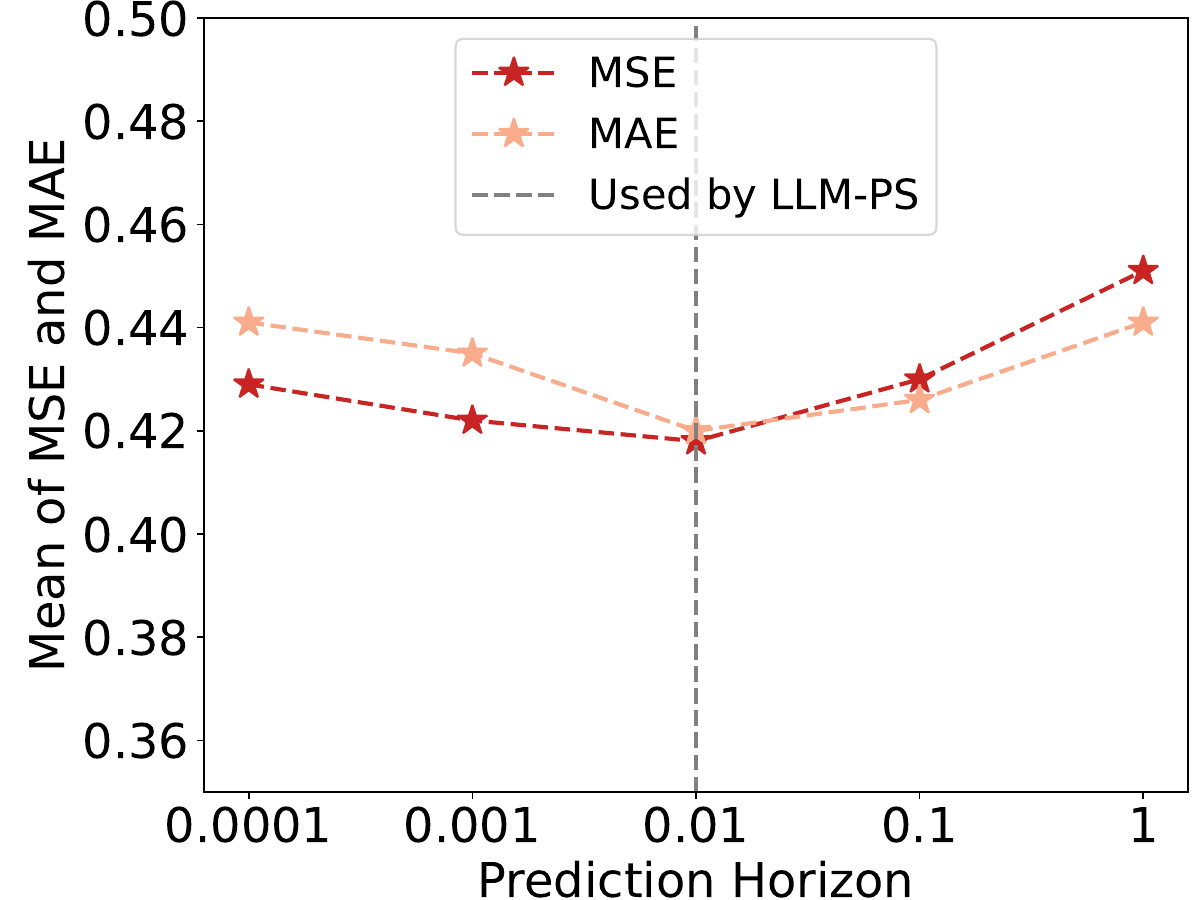}
	\caption{Parametric sensitivities of $\lambda$ in Eq.~\ref{eq_total}.}
	\label{fig_ps}
	\vspace{-7mm}
\end{wrapfigure}
\subsection{Time-to-Text Module}
\label{sec_apx_t2t}
In this paper, we design the Time-to-Text (T2T) module to extract semantic information from the input time series. We follow~\cite{hsu2021hubert} and construct the T2T with an encoder-decoder architecture, where both the encoder and decoder are stacked by transformer layers, as shown in Fig.~\ref{fig_t2t}. \\
\textbf{Model Configuration.} In the T2T module, the encoder and decoder contain 4 and 1 transformer layer, respectively. Each transformer layer consists of the multi-head attention layer, feedforward network, and layer normalization. The patch size, hidden dimension, feedforward hidden size, and output size are set to 24, 96, 384, and 24, respectively. \\
\textbf{Training Setups.} During training, the T2T module is optimized by Adam, where the learning rate is set to 0.001. The masking ratio of input time series patches is set to 0.75.\\
\textbf{Time Series Patching.} The input series $\mathbf{X}=\left\{\mathbf{x}_{1}, \ldots, \mathbf{x}_{H}\right\}\in \mathbb{R}^{H \times V}$ is divided into $P$ overlapping patches $\{\mathbf{X}_{i}\}^{P}_{i=1}$, where $\mathbf{X}_{i}\in \mathbb{R}^{L \times V}$. The patch length is given by $L=\frac{H+O}{P}$, and $O$ denotes the overlap length between consecutive patches.\\
\textbf{Semantic Filtering.} In the LLM word embeddings, there are numerous inrelevant semantic information for Time Series Forecasting (TSF). As shown in Fig.~\ref{fig_t2t}, the original word embeddings $\mathbf{E}_{\text{ori}}\in \mathbb{R}^{W \times D}$ (where $W$ and $D$ are the length and dimension of $\mathbf{E}_{\text{ori}}$) of the GPT2 model includes embedding with corresponding word such as ``Thro", ``Magn", and ``belt", which are irrelevant to TSF. To facilitate T2T in extracting valuable semantic information for TSF, we filter the LLM word embeddings based on text indices relevant to TSF.

Specifically, given the text indices $\{t_{i}\}_{i}^{I}$ ($I$ denotes the number of text indices) corresponding to the TSF, the similarities between their word embeddings $\hat{\mathbf{E}}=\{\mathbf{E}_{\text{ori}}[t_{i}]\}_{i=1}^{I}$ and LLM word embeddings $\mathbf{E}_{\text{ori}}$ are computed as follows: 
\begin{equation}
	\mathbf{s}=\frac{\hat{\mathbf{E}} \cdot \mathbf{E}_{\text{ori}}}{\|\hat{\mathbf{E}}\|_2\|\mathbf{E}_{\text{ori}}\|_2}.
\end{equation}
Then, we select the top 100 most similar word embeddings as the final word embeddings $\mathbf{E}$ for T2T training.
\subsection{Parameter Sensitivity Analysis}
\label{sec_apx_ps}
The LLM-PS equation (Eq.~\eqref{eq_total}) in our model has one tuning parameter, denoted as $\lambda$. We analyze its sensitivities on the ETTh1 dataset by varying $\lambda$ within the range of $\{0.0001, 0.001, 0.01, 0.1, 1\}$ and observe the mean MSE/MAE across predict lengths spanning $\{96, 192, 336, 720\}$, as shown in Fig.~\ref{fig_ps}. Despite the large fluctuations in $\lambda$, the MSE/MAE curve of our LLM-PS remains relatively stable. These results demonstrate the robustness of our LLM-PS against parameter variations. Furthermore, our LLM-PS achieves its best performance when $\lambda=0.01$, so we adopt this parameter configuration in our method.

\end{document}